\pdfoutput=1
\documentclass[11pt]{article}

\usepackage[final]{acl}

\usepackage{times}
\usepackage{latexsym}

\usepackage[T1]{fontenc}
\usepackage[many]{tcolorbox}

\usepackage[utf8]{inputenc}

\usepackage{microtype}

\usepackage{inconsolata}

\usepackage{graphicx}

\usepackage{adjustbox}
\usepackage{tabularx}
\usepackage{todonotes}
\usepackage{footnote}
\usepackage{graphicx}
\usepackage{fancyvrb} % for "\Verb" macro
\usepackage{hyperref}
\usepackage{listings}
\usepackage{placeins}
\usepackage{amsmath}
\usepackage{tabularx}
\usepackage{comment}
\usepackage{enumitem}
\setlist[itemize]{noitemsep}

\usepackage[utf8]{inputenc}
\usepackage{booktabs}
\usepackage{caption}
\usepackage{subcaption}
\usepackage{enumitem}
\usepackage{authblk}
\usepackage{caption}
\usepackage{multirow}
\usepackage{wrapfig}
\usepackage{natbib}
\setcitestyle{round}

\title{Data, Data Everywhere:  \\ A Guide for Pretraining Dataset Construction}

\author{
  \textbf{Jupinder Parmar\thanks{Correspondence to: \texttt{jupinderp@nvidia.com}}},
  \textbf{Shrimai Prabhumoye},
  \textbf{Joseph Jennings},
\\
  \textbf{Bo Liu},
  \textbf{Aastha Jhunjhunwala},
  \textbf{Zhilin Wang},
  \textbf{Mostofa Patwary},
  \\
  \textbf{Mohammad Shoeybi },
  \textbf{Bryan Catanzaro }
\\
   NVIDIA
}

\begin{document}
\maketitle

%Despite the pretraining dataset being a crucial component to the success of new language models
\begin{abstract}

The impressive capabilities of recent language models can be largely attributed to the multi-trillion token pretraining datasets that they are trained on. However, model developers fail to disclose their construction methodology which has lead to a lack of open information on how to develop effective pretraining sets. To address this issue, we perform the first systematic study across the entire pipeline of pretraining set construction. First, we run ablations on existing techniques for pretraining set development to identify which methods translate to the largest gains in model accuracy on downstream evaluations. Then, we categorize the most widely used data source, web crawl snapshots, across the attributes of toxicity, quality, type of speech, and domain. Finally, we show how such attribute information can be used to further refine and improve the quality of a pretraining set. These findings constitute an actionable set of steps that practitioners can use to develop high quality pretraining sets. 

%further refine a pretraning set and improve 

%for each step of the pretraining set development process
%We hope that sharing these results will enable further advancements in the development of language models.

%Finally, we demonstrate how to use the stated findings to construct a pretraining set of English corpora totalling over 5 trillion tokens. 
\end{abstract}

\section{Introduction}

Recent language models (LMs) \citep{openai2024gpt4, geminiteam2024gemini, geminiteam2024geminipro1.5, claude32024, rekateam2024reka} have shown very strong capabilities on a number of evaluation areas. In comparison to previously developed LMs \citep{brown2020language, gpt2-radford2019language, smith2022using, rae2022scaling, workshop2023bloom}, these newly released models generally follow the same architectural details, based on the transformer \citep{Vaswani+2017}. Rather, with emphasis being placed on the size and quality of the pretraining dataset \cite{hoffmann2022training, longpre2023pretrainers}, the improved capabilities of LMs are largely due to self-supervised pretraining on ever larger, higher quality datasets. 
It is clear that the pretraining set is crucial to model success, but the question on how to effectively create one has yet to be openly answered.

%create a good pretraining set has yet to be openly answered. 

%have depended upon self-supervised pretraining on ever larger datasets -- now spanning multiple trillions of tokens. It is clear that the pretraining set is crucial to model success, but the question on how to create a good pretraining set has yet to be openly answered. 

Most leading models \citep{openai2024gpt4, geminiteam2024gemini, claude32024, jiang2023mistral} do not divulge what methods were used to go from raw data sources to a final pre-training set. Other models document only small sections of their process \citep{touvron2023llama2, parmar2024nemotron4, bai2023qwen, rekateam2024reka} and lack information on why or how the chosen decisions were made. The scarcity of open knowledge in this area hinders the general community from contributing to the advancement of model capabilities \citep{rogers2021changing}. 

%both occludes harmful 

%certain aspects of their process such as which data curation procedures were used \citep{touvron2023llama2, parmar2024nemotron4, bai2023qwen} or a high-level breakdown of the percentage of the pretraining set associated to different data sources \citep{rekateam2024reka, parmar2024nemotron4}. In all instances, there is a lack of transparency and information on how the chosen decisions affect the model's resulting  or why certain techniques were used over others. 

%Yet, in all instances there is a lack of transparency and information on how the chosen decisions affect the model's resulting  or why certain techniques were used over others. Given the growing role that pretraining sets play in LM development, the lack of open knowledge in this area has hindered progress and advancements that can continue to push model capabilities forward.
%The general steps in the process for constructing pretraining datasets is shown

\begin{figure*}[h]
    \centering
    \includegraphics[width=\linewidth]{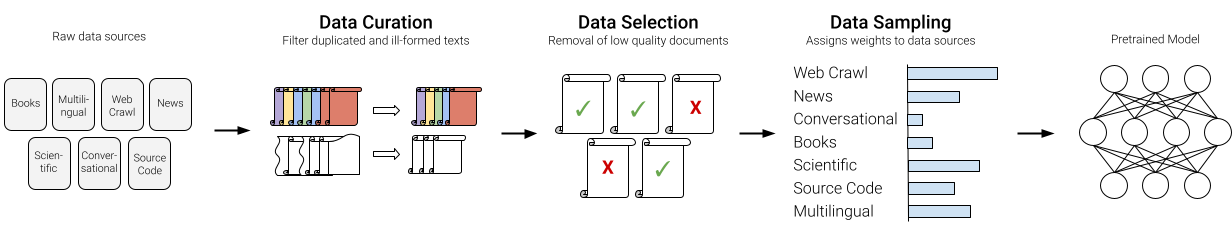}
    \caption{Each step in the development process to go from a collection of data sources into a final pretraining set that produces a highly capable LM.}
    \label{fig:pretraining_pipeline}
\end{figure*}

The steps in pretraining set construction are shown in Figure \ref{fig:pretraining_pipeline}: the pipeline starts with a collection of text data sources, removes ill-formed and duplicate documents during data curation, further filters out low-quality documents via data selection, and finally assigns sampling weights to determine the prevalence of each data source during training. Recent works \citep{longpre2023pretrainers, penedo2023refinedweb, soldaini2024dolma, fine-web} have started to elucidate strategies for effective pretraining set development. However, they all focus solely on the step of data curation and analyze only a small number of mostly English sources. 

%The purpose of the ablations is to take a bunch of the stuff that has been put out there and quantify what works and what doesn’t in the steps of pretraining development

%* 		We first run a series of ab- 069 lations at each step of the development pipeline 070 to determine which methods help create the most 071 performant pretraining
%* 		Is there a way to make this concrete?
%* 		like what are these ablations? ... maybe not call this ablation ... more like comparison?
%* 		the why can also be more concrete - instrad on "help create most ..."
%* 		should be something like to understand / insight into X
%* 		say "at each step of the development pipeline in Fig 1"
%We first run a series of ablations at each step of the development pipeline to determine which methods help create the most performant pretraining set. 

%For the best identifeid method we 

In this paper, we provide insights across all steps of pretraining set development for a set of over 2T tokens composed of English, multilingual, and source code documents. We compare existing methods through a series of ablations at each step of the development pipeline in Figure \ref{fig:pretraining_pipeline} to quantify which techniques do and do not realize improvements in downstream evaluations. For the best identified method, we highlight various design decisions that impact performance. 

Additionally, previous studies on web crawl are conducted across a small number of snapshots and are limited to the characteristics of toxicity and quality \citep{longpre2023pretrainers}. Despite web crawl documents constituting the majority of examples in pretraining sets \citep{almazrouei2023falcon, smith2022using, gemma24}, we still do not thoroughly understand their composition. We close this gap by conducting a large-scale analysis on over 90 Common Crawl web snapshots for the attributes of domain, quality, toxicity, and type of speech. We then show how such data attributes can aid in pretraining set construction to further improve model capabilities.

By sharing this information, we provide an actionable series of steps that can be used to construct highly performant pretraining sets. Concretely, our contributions are as follows:

\begin{enumerate}[topsep=0pt,itemsep=2pt,partopsep=1ex,parsep=1ex]
\item[\textbullet] Suggest a set of techniques to use for the data curation, selection, and sampling steps of pretraining set development for English, multilingual, and code data.
  \item[\textbullet] Perform the first large-scale analysis of web crawled data across the attributes of quality, toxicity, type of speech, and domain.
  \item[\textbullet] Demonstrate that attribute information can be used to enhance the performance of data sampling and data selection methods.
\end{enumerate}

\section{Experimental Setup}

%We identify how 

We ablate a singular part of the development pipeline and train a LM on the resulting pretraining set to understand how various methods affect performance on downstream benchmarks. Our experimental setup is detailed below.

%We explore how various design decisions in this process affect performance on downstream benchmarks by ablating a singular part of the pipeline and training a LM on the resulting pretraining set. The experimental setup and axes across which we experiment are detailed below with results following in later sections. 

%We illustrate the nuanced process required to construct pretraining datasets in Figure 1: the pipeline starts with a collection of text data sources, removes low-quality or duplicate documents during data curation, further filters out unessential documents via data selection, and finally applies sampling weights to determine the prevalence of each data source during training. We explore how various design decisions in this process affect performance on downstream benchmarks by ablating a singular part of the pipeline and training a LM on the resulting pretraining set. The experimental setup and axes across which we experiment are detailed below with results following in later sections. 

\subsection{Data Sources}

With current language models being trained on a wide range of data sources, an appropriate study on pretraining set construction must use a large, diverse set of data. Table \ref{tab:data_high_level} highlights the sources, along with the amount of tokens coming from each, included in the English, multilingual, and code data that we use in our experiments.

%The data in our English sources come from publicly available datasets like The Pile \citep{pile-dataset-2020}, Common Crawl (CC), BigScience ROOTS \citep{lachaux2020unsupervised} and more; for a full breakdown please refer to Appendix \ref{sec:appendix_data_sources_english} and Table \ref{tab:english_data_summary}. Our multilingual dataset consists of monolingual data in 52 languages as well as an English-centric sentence-level parallel corpus; for a full breakdown please refer to Appendix \ref{sec:appendix_data_sources_multilingual} and Tables \ref{tab:ml_data_summary} and \ref{tab:nmt_composition}. Our source code dataset contains 41 languages and was obtained from StarCoder \cite{li2023starcoder}, which in turn is derived from the Stack\citep{kocetkov2022stack}; for a full breakdown please refer to Appendix \ref{sec:appendix_data_sources_code} and Table \ref{tab:code_data_summary}.

%Given the diverse types of data that are used to train language models today, we felt that to conduct an appropriate study on pretraining set construction our experiments must incorporate a  wide-ranging set of data sources. Table \ref{tab:data_high_level} highlights the sources, along with the amount of tokens coming from each, included in the English, multilingual, and code data that we use in our experiments. 

\begin{table}[!h]
\centering
  \begin{tabular}{llr}
   \toprule
    \textbf{Data type} & \textbf{Data source} & \textbf{Tokens (B)} \\
    \midrule
    \multirow{6}{*}{English} & 
    Web crawl &  889 \\
    & Misc &  109   \\
    &  News   & 94  \\ 
     & Conversational  & 59    \\
     & Books &  35 \\ 
    & Scientific   & 33 \\
    \midrule
    \multirow{2}{*}{Multilingual} \hspace*{-0.25cm}
    & Web crawl   & 540  \\
    & Parallel corpora  & 56  \\
    \midrule 
    \multirow{1}{*}{Source Code} &  The Stack v1.2 & 212  \\
    \bottomrule
  \end{tabular}
  \caption{The data sources that are used in our ablation studies. Table \ref{tab:english_data_summary}, Table \ref{tab:ml_data_summary}, Table \ref{tab:nmt_composition}, and Table \ref{tab:code_data_summary} provide a more detailed breakdown of the English, multilingual, and source code datasets.}
  \label{tab:data_high_level}
\end{table}

%A more comprehensive breakdown of the English data sources are in Table \ref{tab:english_data_summary}, multilingual in Table \ref{tab:ml_data_summary}, parallel corpora in Table \ref{tab:nmt_composition} and source code in \ref{tab:code_data_summary}.

%% also maybe don't need to 

Experimenting on this broad set of data ensures our findings will be applicable in the development of large-scale pretraining sets. As current language models do not just pretrain on English-only data, we highlight the importance of including multilingual and code data within our study. However, while we run ablations for these domains, the majority of our experiments focus on the English set.

\subsection{Evaluation}
In experiments on English datasets, we use the LM-Evaluation Harness \citep{gao2021empirical} to evaluate zero-shot accuracy on  PIQA \citep{Bisk2020PIQARA}, ARC-easy \citep{clark2018think}, Winogrande \citep{Sakaguchi2020WINOGRANDEAA}, Hellaswag \citep{Zellers2019HellaSwagCA}, LAMBADA \citep{paperno-etal-2016-lambada}, and Race-H \citep{lai-etal-2017-race}. We also evaluate on MMLU \citep{hendrycks2020measuring} when the experimental setting allows for a non-random score. For our source code experiments, we evaluate on HumanEval and  MultiPL-E \citep{chen2021evaluating, multiple2023}. In our multilingual experiments, we evaluate on XCOPA \citep{xcopa} and TyDiQA-GoldP \citep{tydiqa}.

%along with the code translation task TransCoder as introduced in \citep{chowdhery2022palm}.

\subsection{Model Specifications}
To ensure that our results hold at various model scales, in our experiments we use either 2B or 8B decoder only transformer LMs trained with autoregressive language modeling at token horizons from 150B to 450B tokens. The configuration used for a given experiment is specified ahead of each reported result. Specifics on model architecture and hyperparameters are shared in Appendix \ref{sec:appendix_model_specifications}. 

% We train all models using autoregressive language modeling with the cross entropy loss over the prediction of the next token. 

%The configuration that is used for a given experiment will be specified ahead of each reported result.

\section{Data Curation}

\subsection{Methodology}

As dataset curation has been widely investigated, we do not run ablations to identify which specific techniques are beneficial, but rather compare the benefit when using these studied techniques versus not. We consider three phases of data curation: raw text, post deduplication, and post quality filtering. Our deduplication process is comprised of both exact deduplication where we compute a 128-bit hash for each document, group the documents by their hashes, and select one document per group in addition to fuzzy deduplication as described in \citep{smith2022}. In quality filtering, the deduplicated documents are filtered based on the perplexity of a KenLM model \citep{heafield-2011-kenlm}  that was trained on a collection of high quality sources alongside a set of heuristic filters as described in \citep{Rae2021Gopher, raffel2020exploring}. Full details on the quality filtering steps are shared in Table \ref{tab:filtering_heuristics}. 
When curating the source code datasets we formed repository-level contexts and filtered out low-quality documents by following the approach of \cite{li2023starcoder}, which is outlined in Table \ref{tab:code_filtering_heuristics}.

\subsection{Ablations}

\begin{tcolorbox}[breakable,width=\linewidth, colback=white!95!black, title={Findings}]
\begin{itemize}
  \item Compared to raw text, deduplicated and quality filtered data improve model accuracy.
  \item In deduplication, it is better to priortize keeping samples from older sources than more recent ones.
\end{itemize}

\end{tcolorbox}

All our data curation experiments use a 2B parameter model trained for 300B tokens. Table \ref{tab:curation_results} shows that model accuracy improves after both deduplication and quality filtering, indicating the utility of effective data curation. The impact of data curation for code is shared in Appendix \ref{sec:appendix_data_curation_ablations}. 

\begin{table}[!h]
\centering
  \begin{tabular}{lr}
   \toprule
    \textbf{Experiment} & \textbf{LM-Eval} \\
    \midrule
    Raw text &  57.18 \\
    Post deduplication   & 58.93  \\ 
    Post quality filtering  & \textbf{59.50}    \\
    \bottomrule
  \end{tabular}
  \caption{Impact of data curation steps on model accuracy. Per-task accuracies are shared in Table \ref{tab:full_data_curation_results}.}
  \label{tab:curation_results}
\end{table}

%In fuzzy deduplication it is possible to prioritize documents from certain sources. To determine how this priortization would be best defined, we run a series of ablations. 

%We run ablations to determine how this priortization should be defined. 

In fuzzy deduplication, it is possible to priortize retaining documents from certain sources. As document age has been shown to impact model accuracies  \citep{longpre2023pretrainers}, we run ablations with the following priortization of data sources: most recent to oldest, oldest to most recent, or at random. Table \ref{tab:data_dedup_results} indicates that prioritizing older documents leads to significantly better results.

%This result confirms previous findings that when a similar document is contained in both an older and more recent dataset, the text in the more recent document likely contains higher quantities of emojis and non-ASCII punctuation \citep{longpre2023pretrainers} which degrade its quality.

\begin{table}[!h]
\centering
  \begin{tabular}{lr}
   \toprule
    \textbf{Experiment} & \textbf{LM-Eval} \\
    \midrule
    Random &  59.96 \\
    Recent-to-Old   & 58.93  \\ 
    Old-to-Recent  & \textbf{60.47}    \\
    \bottomrule
  \end{tabular}
  \caption{The priortization of data sources in deduplication affects model accuracy. Per-task accuracies are shared in Table \ref{tab:full_data_dedup_results}.}
  \label{tab:data_dedup_results}
\end{table}

\section{Data Selection}

\subsection{Methodology}
In addition to filtering done during data curation, specialized methods have been developed for data selection \citep{albalak2024survey} to ensure that only the highest quality documents make it into pretraining corpora. Amongst the potential methods, we specifically investigate and run ablations with Domain Selection via Importance Resampling (DSIR) \citep{xie2023data} as it requires minimal compute overhead and is part of the set of techniques that stem from Moore-Lewis selection \citep{moore-lewis-2010-intelligent}, which accounts for most data selection methods. DSIR takes as input a raw dataset, along with a target dataset of known high quality examples, and then uses importance resampling to select examples from the raw dataset that are distributed like the target by utilizing a bag of hashed n-gram models to match the n-gram frequencies of the selected data and the target.

\subsection{Ablations}
\label{sec:selection_ablation}
\begin{tcolorbox}[breakable,width=\linewidth, colback=white!95!black, title={Findings}]
\begin{itemize}
\item DSIR improves the quality of web crawl snapshots.
\item DSIR functions best when applied across each data source individually.
\item DSIR is fairly sensitive to the composition of the target distribution.
\end{itemize}
\end{tcolorbox}

We assess whether DSIR provides gains when used on data that has passed through a data curation pipeline. Through our ablations, we seek to answer: 1) how does naive application with the recommended settings of DSIR perform and 2) can we identify better settings for DSIR. In tackling question 2, we ablate whether selection should be done at the level of individual data sources instead of the entire pretraining corpus and altering the suggested percentage of data that should be selected. All our DSIR experiments train a 2B parameter model for 165B tokens on a training set of two CC snapshots.

\begin{table}[!h]
\centering
  \begin{tabular}{llc}
   \toprule
    \textbf{Question} & \textbf{Experiment} & \textbf{LM-Eval} \\
    \midrule
    \multirow{2}{*}{Q1} & 
    Original CC &  54.30 \\
    &  DSIR    & \textbf{54.44}  \\ 
     \midrule
    \multirow{2}{*}{Q2.1} \hspace*{-0.25cm}
    & Corpus DSIR   & 54.44  \\
    & Source DSIR & \textbf{54.71}  \\
    \midrule
    \multirow{3}{*}{Q2.2} \hspace*{-0.25cm}
    & DSIR (80\%)   & 54.55  \\
    & DSIR (87.5\%)   & 54.25  \\
    & DSIR (95\%)  & \textbf{54.71}  \\
    \bottomrule
  \end{tabular}
  \caption{DSIR improves the quality of web crawl data. () refers to the percentage of examples that are selected by DSIR. Per-task accuracies are shared in Table \ref{tab:full_dsir_main_results}.}
  \label{tab:dsir_main}
\end{table}

As shown in Table \ref{tab:dsir_main}, the naive application of DSIR, using a target set of Wikipedia and Books, leads to a slight improvement in accuracy compared to post curation CC data, 54.48 vs 54.30. We find that selecting at the level of individual sources improves upon the paper-recommended setting of selection across the entire corpus. The recommended 95\% selection rate is optimal.

We ran an additional ablation to understand the sensitivity in performance of DSIR when the target set is altered. Table \ref{tab:dsir_target_set_ablation} illustrates that even small alterations to the target set, such as the addition of a high quality source like arXiv, causes fluctuations in model accuracy  -- indicating that the target set should be defined carefully.

%Even Small alterations to the target set recommended by the paper, Wikipedia and books, such as the addition of a high quality source like arXiv, causes significant -- indicating that the composition of the target set should be defined carefully. fluctuations in model performance as shown in Table \ref{tab:dsir_target_set_ablation}

%Additionally, we confirm that 

%Table \ref{tab:dsir_main} highlights that DSIR does lead to performance improvements as the post curation CC data obtains an average accuracy of 54.30 compared to 54.71 for DSIR. In the ablations which explore the optimal setting for DSIR, we find that the paper specified selection rate of 95\% is ideal and that it is best to apply DSIR on a per data source basis. 

%These experiments all used the target set recommend by the paper, Wikipedia and books. We ran an additional ablation to understand the sensitivity in performance when the target set is altered. As shown in Table \ref{tab:dsir_target_set_ablation} we find that even small alterations to the content of information contained in the target set, such as the addition of a high quality source like arXiv, causes significant fluctuations in model performance -- indicating that the composition of the target set should be defined carefully. 

%Fully detailed results of this ablation can be found in Appendix \ref{sec:appendix_data_selection_ablations} and Table \ref{tab:dsir_target_set_ablation}. 

\begin{table}[!h]
\centering
  \begin{tabular}{lr}
   \toprule
    \textbf{Target Set} & \textbf{LM-Eval} \\
    \midrule
    Wikipedia, Books &  \textbf{54.71}  \\
    Wikipedia, Books, arXiv, NIH   & 54.02  \\ 
    arXiv, NIH   & 53.71  \\ 
    \bottomrule
  \end{tabular}
  \caption{DSIR is impacted by target set composition. Per-task accuracies are shared in Table \ref{tab:full_dsir_target_set_ablation}.}
  \label{tab:dsir_target_set_ablation}
\end{table}

\section{Data Sampling}

\subsection{Methodology}

During the construction of pretraining corpora, data weights $\{a_k\}^{N}_{k = 1} \text{ such that } \sum_{k =1}^{N} a_k = 1$ are assigned to each of the $N$ data sources to determine the sampling frequency of each source during pretraining. The value of data weights can greatly impact downstream accuracy as increasing the proportion of data from a given source decreases the cumulative weight on the others, potentially causing degradation on the domains that are now less represented. Specialized methods have been developed to identify appropriate sampling weights that endow the trained model with strong capabilities across a wide range of domains.

In our ablations, we consider two data sampling methods that use heuristics based on characteristics of the data sources to define weight distributions: alpha sampling \citep{arivazhagan2019massively, shliazhko2022mgpt} and UniMax sampling \citep{chung2023unimax}, in addition to DoReMi \citep{xie2023doremi} which uses a learned model to identify the sampling weights. Both alpha and UniMax sampling use the number of tokens in each data source to define data weights. Alpha sampling proportionally weights data sources to a scaled factor, $\alpha$, of their token counts while UniMax fits a uniform weight distribution subject to the constraint that no data sources sees more than a 
certain number of epochs at the given training token budget. Comparatively, DoReMi defines data weights by formulating the problem via group distributionally robust optimization \citep{sagawa2020distributionally} and minimizing the excess loss between a small proxy model and a pretrained reference model.

\subsection{Ablations}

\begin{tcolorbox}[breakable,width=\linewidth, colback=white!95!black, title={Findings}]
\begin{itemize}
\item UniMax provides the best sampling weights for the English and multilingual domains. 
\item Alpha sampling, with a value of  $\alpha=1.3$ , provides the best sampling weights for the code domain.
\item DoReMi is unable to produce competitive sampling weights for any domain as it often gives the majority of the weight to a single source.
\end{itemize}
\end{tcolorbox}
%Experiments from doc + results + analysis

%The maximum epoch hyperparameter in UniMax should be 

%For the English and multilingual domains, UniMax provides the best sampling weights. The number of maximum epochs in UniMax should be defined based on a reasonable quantity based on the number of training tokens. 

In our data sampling ablations, we study the domains of English, multilingual, and code individually as the inherent characteristics of each domain would likely change which data sampling method would be best suited for it. We use an 8B parameter model for the ablations and train on 150B tokens for the code domain and 300B tokens for the English and multilingual domains. 

%The best result for each method from these ablations is detailed below and the full set of results can be found in Appendix \ref{sec:appendix_data_sampling_methods}. 

\subsubsection{English}

In our English ablations, we replace alpha sampling with preference based weighting, where the weights are hand tuned according to intuitive perceptions of quality, as it has been the most widely used sampling technique for English data \citep{touvron2023llama, gao2020pile}. With the weights returned by Unimax being dependent upon the number of epochs allowed for each data source, we additionally ablate across across varying values of this hyperparameter. The returned sampling weights and further details on each method can be found in Appendix \ref{sec:appendix_data_sampling_ablations}.

%As the weights returned by UniMax are dependent upon the number of epochs that a data source can maximally be used for, we provide results across varying values of this hyperparameter to demonstrate its impact on performance.  The exact setting and returned sampling weights for each method can be found in Appendix \ref{sec:appendix_data_sampling_ablations}. 

%For our English ablations, in place of alpha sampling we use preference based weighting, where the practitioner hand tunes weights according to intuitive perceptions of quality, as it has been the most well studied and reported in literature for English data \citep{touvron2023llama, gao2020pile}. 

\begin{table}[!h]
\centering
  \begin{tabular}{lcc}
   \toprule
    \textbf{Method} & \textbf{LM-Eval} & \textbf{MMLU} \\
    \midrule
  
    Preference &  65.85 & 27.20 \\
      UniMax (1e)   & \textbf{67.14} & \textbf{28.30}  \\ 
      UniMax (2e)   & 66.50 & 28.00  \\ 
      UniMax (4e)   & 66.61 &  26.60 \\ 
      DoReMi   & 65.63 & 26.90  \\ 
    %\midrule
    %\multirow{2}{*}{450B} & 
     %Preference   & 65.95  & 27.5 \\ 
    %&  UniMax   & \textbf{67.79}  & \textbf{29.4} \\ 
    \bottomrule
  \end{tabular}
  \caption{UniMax sampling weights provide the best performance on English data. $N$e means that UniMax can use a maximum of $N$ epochs per dataset. Per-task accuracies are shared in in Table \ref{tab:full_data_samp_eng}.}
  \label{tab:data_samp_eng}
\end{table}

Table \ref{tab:data_samp_eng} shows that UniMax achieves substantially better accuracies on LM-Eval and MMLU compared to the next best method. We note that DoReMi attains the worst performance, which we believe to be a factor of its weight distribution being heavily skewed towards web crawl snapshots as detailed in Appendix \ref{sec:appendix_data_sampling_ablations}. Additionally, despite still outperforming both other methods, the performance of UniMax degrades as the maximum epoch hyperparameter increases. We hypothesize that as we have far more data tokens than the amount of training tokens, repeated epochs of data provide less utility than novel information. We suggest that practitioners choose the minimal value of the epoch hyperparameter that makes sense for their datasets and training budget.

%mitigates choose the minimal value for the epoch hyperparameter which does not result in a purely proportional sampling distribution. 

\subsubsection{Multilingual}

%Sampling weights for langauges are a lot more complex. We want to blalance

%Makes sense that unimax does good here as helps prevent overfitting to langauge and also language exhibit transfer cite so atually the absoltue weight

%alpha sampling, unimax, doremi 

%300B tokens

% alpha sampling alpha stuff in appendix

%Makes sense that unimax does good here as helps prevent overfitting to langauge and also language exhibit transfer cite so atually the absoltue weight

%Table \ref{tab:data_samp_mult}. highlights that on multilingual data, unimax sligtly pefroma alpha sampling. Additionally, the weights provided by DoReMi have a isngi

It has been shown that models trained on a subset of multilingual languages from a given language family are able to transfer knowledge and capabilities to other languages in the family \citep{k2020crosslingual, hu2020xtreme, ye2023language}. This indicates that a sampling method which more evenly spreads weight so that all language families are well represented, like UniMax, should achieve better accuracy than one which places most of the weight on high resource languages, like alpha sampling. Table \ref{tab:data_samp_mult} confirms this intuition as UniMax slightly outperforms alpha sampling. As with the English ablations, DoReMi's returned weight distribution is heavily skewed, causing it to underperform both other methods. The sampling weights identified by each method are detailed in Appendix \ref{sec:appendix_data_sampling_ablations}.

\begin{table}[!h]
\centering
  \begin{tabular}{lcc}
   \toprule
      \textbf{Method} & \textbf{XCOPA} & \textbf{TyDiQA-GoldP} \\
    \midrule
    Alpha ($\alpha$ = 1.3)   &  58.11 & 17.86 \\
     UniMax (1e)   & \textbf{58.24} & \textbf{18.11}  \\ 
      DoReMi   & 57.65 & 15.8 \\ 
    \bottomrule
  \end{tabular}
  \caption{UniMax slightly outperforms alpha sampling on multilingual data.}
  \label{tab:data_samp_mult}
\end{table}

%Per-language accuracies are shared in Table \ref{tab:full_data_samp_mult}.

\subsubsection{Code}

We do not use the returned DoReMi sampling distribution in our code ablations as it placed over 80\% of the weight on a single programming language, which does not allocate enough tokens to facilitate model learning during training for the remaining 42 languages. As shown in Table \ref{tab:data_samp_code}, we find that alpha sampling achieves better accuracies than UniMax. In our study, we did not find there to be a strong transfer ability between programming languages as has been seen for multilingual languages. Given that we mainly evaluate on high resource languages, we find it natural that alpha sampling, which places high weight on high resource languages without dramatically undersampling low resource languages, performs best. Further details on this ablation can be found in Appendix \ref{sec:appendix_data_sampling_ablations}.

%During our study, we consistently saw that the ordering of sampling weights for programming languages nearly directly matched the ranking of the model's evaluation performance on the languages. In other words, there is not a strong transfer ability between programming languages as has been seen for multilingual languages.

%All details on this ablation can be found in Appendix \ref{sec:appendix_data_sampling_ablations}

\begin{table}[!h]
\centering
  \begin{tabular}{lcc}
   \toprule
      \textbf{Method} & \textbf{MultiPL-E} & \textbf{HumanEval} \\
    \midrule
    Alpha ($\alpha$ = 1.3)   &  \textbf{19.72} & \textbf{20.73} \\
     UniMax (1e)   & 19.33 & 20.12  \\ 
    \bottomrule
  \end{tabular}
  \caption{Alpha sampling outperforms UniMax on code data. Per-language accuracies for MultiPL-E are shared in Table \ref{tab:full_data_samp_code}.}
  \label{tab:data_samp_code}
\end{table}

\section{Data Attributes} 

\subsection{Methodology}
We investigate attributes along the axes of toxicity, quality, domain, and type of speech for each document that comes from  CC snapshots. Information from quality and toxicity labels can be used to categorize the potential utility of a given document while domain and type of speech labels characterize the types of documents that compose our pretraining set. We obtain these attribute labels by training a DeBERTaV3 \citep{liu2019roberta} classifier on a small set of ground-truth labeled web crawled documents before obtaining predictions from each across our entire pretraining corpus. A full breakdown of the labels that each classifier outputs along with a more detailed description of the classifier training procedure can be found in Appendix \ref{sec:appendix_data_attribute_classifiers}. 

\subsection{Attribute Analysis}
\begin{tcolorbox}[breakable,width=\linewidth, colback=white!95!black, title={Findings}]

\begin{itemize}
\item Website homepages, news articles, and blogs constitute the majority of web crawl documents. Conversational texts are sparsely contained.

\item Technical domains like finance, law, and science are among the least represented in web crawl.

\item Explanatory or news articles on science and health are the most likely to be high quality documents.

\item Domains or types of speech that are generally of high quality may also exhibit high toxicity (i.e news articles on sensitive topics), explaining why previous toxicity based filtering has harmed model accuracy.

\end{itemize}
\end{tcolorbox}

%* 		to improve the quality of pretraining data by identifying places we lack
%This analysis aids pretraining set development by identifying areas of data shortage.

We perform the first large-scale study of web crawl snapshots by using our aforementioned attribute classifiers to analyze all available CC snapshots until August 2023, over 90 in total. This analysis provides new insights into the composition of web crawl documents and identifies areas of data shortage, both of which can be used to improve the quality of pretraining sets. We detail our key findings below and further analysis can be found in Appendix \ref{sec:appendix_data_attribute_analysis}.

%With web crawl snapshots making up a large proportion of pretraining sets, it is important to fully understand the composition and characteristics of these documents.We perform the first such large-scale study by using our aforementioned attribute classifiers to analyze all available CC snapshots until August 2023, over 90 in total.We detail our key findings below and the further analysis can be found in Appendix \ref{sec:appendix_data_attribute_analysis}.

%To do so, we consider all available CC snapshots until August 2023, which is over 90 in total. Using the aforementioned attribute classifiers, we obtain predictions for each document after the snapshots have passed through data curation and analyze the resulting composition.

\begin{figure}[h]
    \centering
    \includegraphics[width=\linewidth]{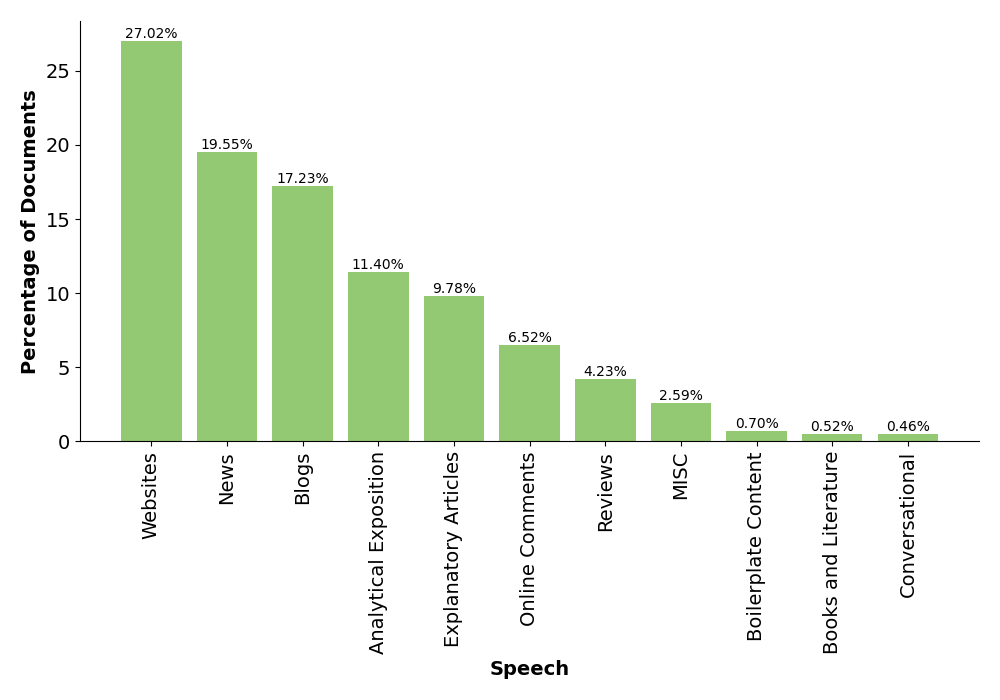}
    \caption{Distribution of document types in web crawl.}
    \label{fig:attribute_speech}
\end{figure}

%Figure \ref{fig:attribute_speech} quantifies the proportion of documents belonging to various types of speech categories. Three major document types make up over 65\% of all web crawl examples: websites (homepages for organizations, products, and individuals), news articles, and blogs. This potentially explains the vastly improved world knowledge of recent LMs \citep{touvron2023llama2, jiang2023mistral} as news and blogs contain information on a wide range of topics while homepages provide factual information on people, places, and items. The lack of conversational texts highlights why alignment is needed to greatly improve the chat ability of pretrained models.

Figure \ref{fig:attribute_speech} quantifies the proportion of documents belonging to various types of speech. Three major document types constitute over 65\% of all web crawl examples: websites (homepages for organizations, products, and individuals), news articles, and blogs. This potentially explains the vastly improved world knowledge of recent LMs \citep{touvron2023llama2, jiang2023mistral} as news and blogs contain information on a wide range of topics while homepages provide factual information on people, places, and items. The lack of conversational texts highlights why alignment is needed to greatly improve the chat ability of pretrained models.

%stuff about books here

\begin{figure}[h]
    \centering
    \includegraphics[width=\linewidth]{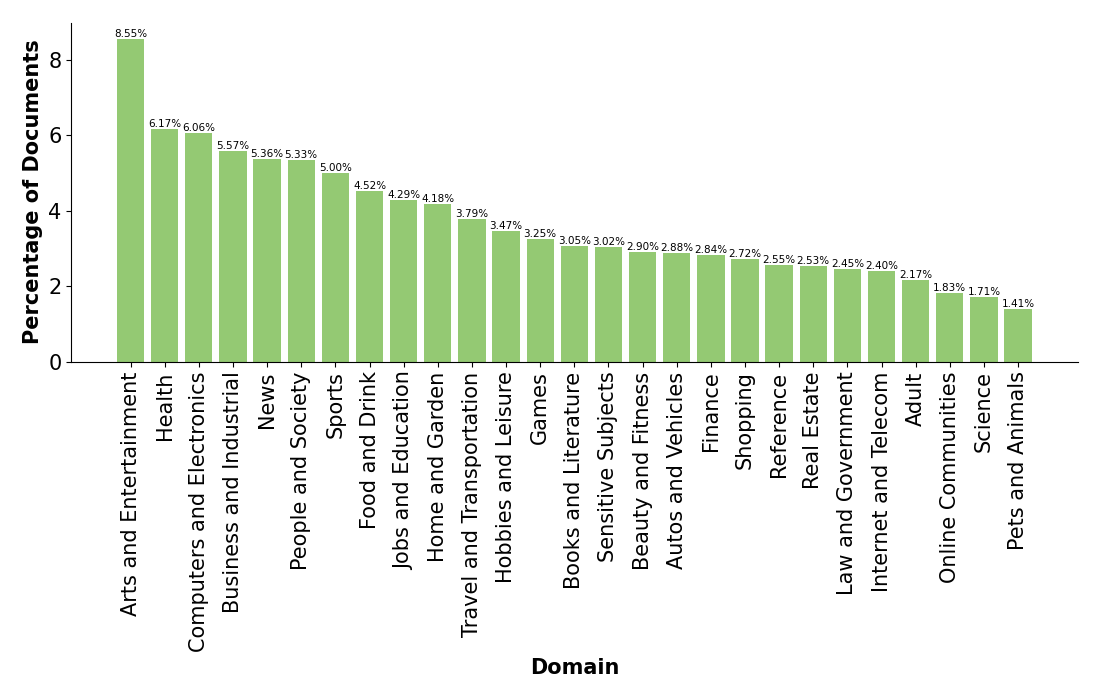}
    \caption{Distribution of content domains in web crawl.}
    \label{fig:attribute_domain}
\end{figure}

Figure \ref{fig:attribute_domain} illustrates the composition of content domains. The domains which are present in lower quantities are often technical in nature: finance, law, and science. To ensure that the model attains strong capabilities in these areas, it is pertinent to augment pretraining sets with data sources such as SEC filings \citep{wu2023bloomberggpt}, Court Listener \cite{henderson2022pile}, and academic papers \citep{gao2020pile, touvron2023llama2}.

%\begin{figure*}
%\centering
%\begin{subfigure}{.5\textwidth}
%  \centering
%  \includegraphics[width=\linewidth]{acl-style-files/figures/8T_Visualizations/Speech_percentage_plot_no_border.png}
%  \label{fig:sub1}
%\end{subfigure}%
%\begin{subfigure}{.5\textwidth}
%  \centering
%  \includegraphics[width=\linewidth]{acl-style-files/figures/8T_Visualizations/Domain_percentage_plot_no_border.png}
%  \label{fig:sub2}
%\end{subfigure}
%\caption{test}
%\label{fig:analysis_tos_domain}
%\end{figure*}

\begin{figure}[h]
    \centering
    \includegraphics[width=\linewidth]{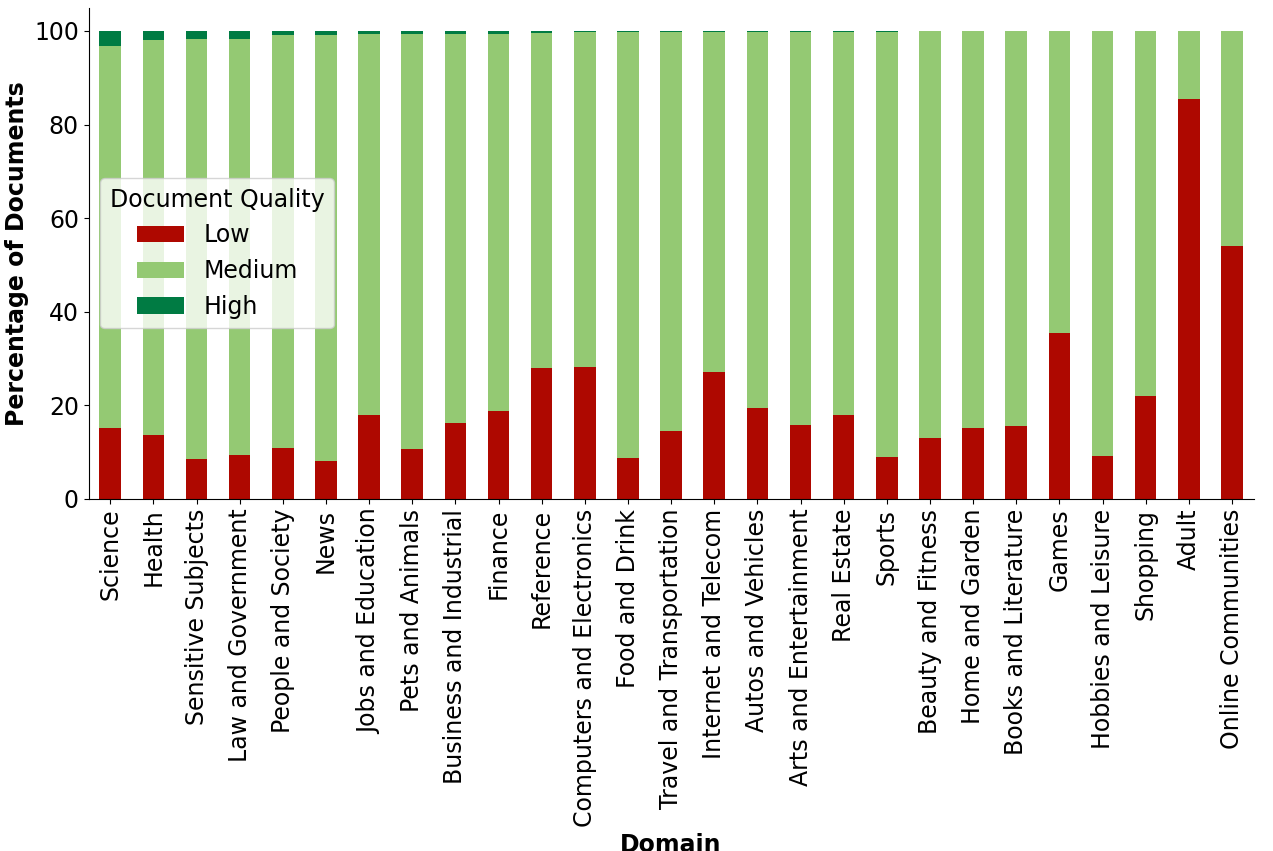}
    \caption{Domains sorted by descending order of percentage of high quality documents.}
    \label{fig:domain_quality}
\end{figure}

We now examine how multiple data attributes vary with each other. Figure \ref{fig:domain_quality} shows the quality composition of each domain. 
As expected, technical domains like science, health and law contain the largest proportion of high quality content while adult and online communities are primarily of low quality. Surprisingly, sensitive subjects contains the third highest percentage of high quality examples. Looking at the distribution of domain by type of speech,  which is detailed in Appendix \ref{sec:appendix_data_attribute_analysis}, the majority of sensitive subjects documents are news articles -- indicating that these are well-written reports on topics such as war and protests.

\begin{figure}[h]
    \centering
    \includegraphics[width=\linewidth]{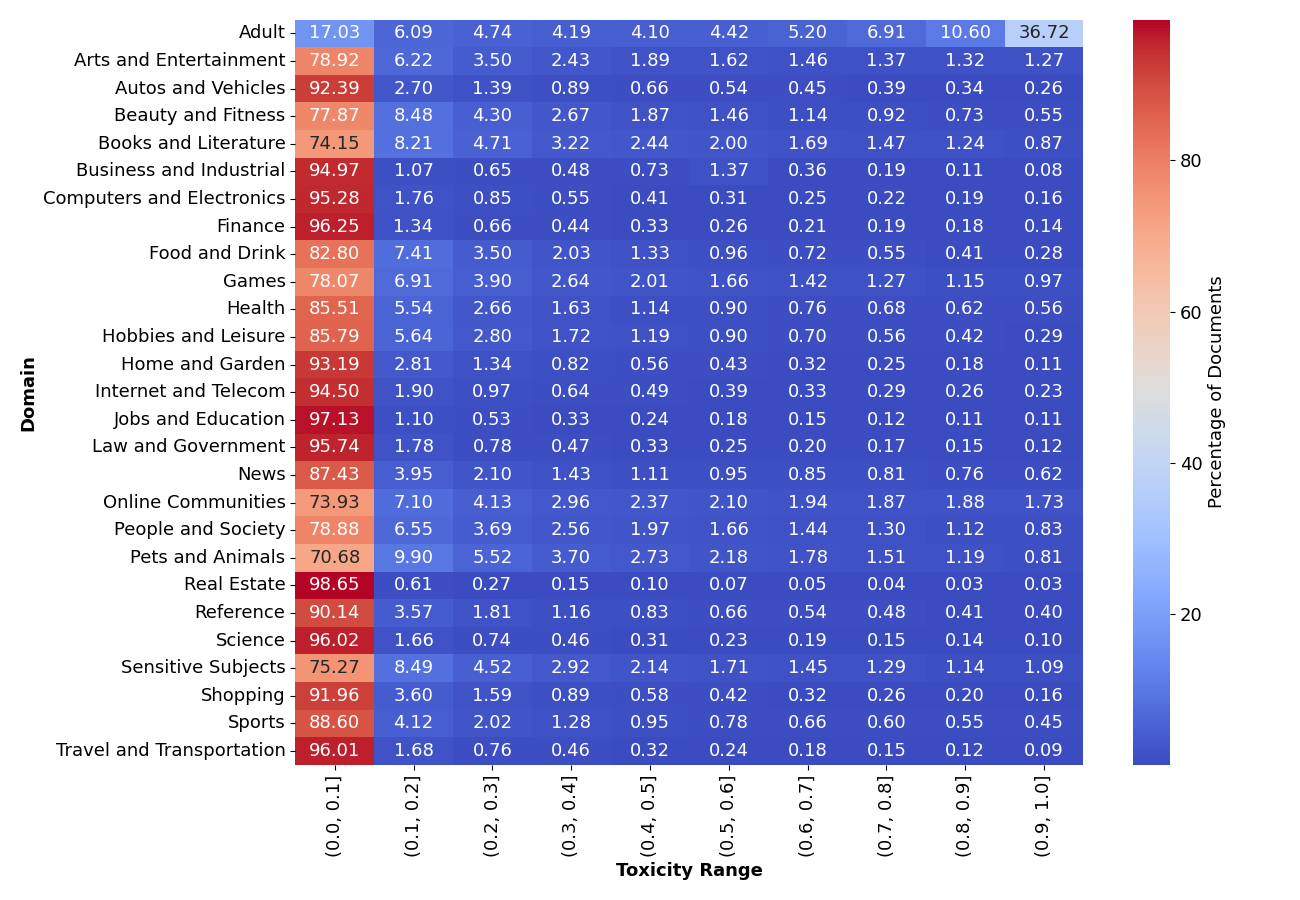}
    \caption{Heatmap of domains by probability of toxic content. Adult and online communities contain the highest percentage of toxic content.}
    \label{fig:domain_toxicity}
\end{figure}

Figure \ref{fig:domain_toxicity} shows the relationship between domain and toxicity. Sensitive subjects, likely due to the contained topics, is flagged for having high toxicity. This illustrates how toxicity based filtering can remove high quality documents and degrade LM quality as shown previously \citep{xu2021detoxifying}.

\subsection{Attributes in Sampling and Selection}
\label{attributes_in_sampling}
\begin{tcolorbox}[breakable,width=\linewidth, colback=white!95!black, title={Findings}]
\begin{itemize}
\item Buckets defined by data attributes substantially improve the performance of data sampling methods. 
\item Attributes compose more useful target sets for data selection.
\end{itemize}
\end{tcolorbox}

Data attributes can refine pretraining set development as more exact target sets can be used in data selection and more informative buckets of data can be defined for which to assign weight distributions over during data sampling. We quantify the benefit of incorporating data attributes within both of these steps.

%The above analysis of data attributes across web crawl documents can aid in the pretraining set development process. With more exact information on the characteristics of a given example, data selection will be more informed and better buckets of data can be defined for which to assign weight distributions over during data sampling. We quantify the substantial added benefit of incorporating data attributes within both of these steps. 

To use attribute information within data sampling, we define new buckets of examples based on the attributes. In one setting, which we term \texttt{fine-grained}, each existing data source is partitioned based on the attribute. A given CC snapshot CC-1 will now become $\{\text{CC-1-}X_i\}_{i = 1}^{n}$ where each $X_i$ is one of the $n$ classes for the attribute. This means $\bigcup\limits_{i=1}^{n} \text{CC-1-}X_i = \text{CC-1}$. An alternative setting, termed \texttt{grouped}, is to create attribute buckets across the entire corpus, $C$, such that $\bigcup\limits_{i=1}^{n} X_i = C$, as each $X_i$ consists of samples among all data sources with that given attribute label. 

%disregard the existing data sources and 

%To incorporate data attribute information within data sampling, we define new buckets of examples based on the attributes. In one setting, which we term \texttt{fine-grained}, each existing data sources is 
%partitioned further based on the attribute.

%Thus, a given CC snapshot CC-1 will now become: CC-1-$X_1$, ... , CC-1--$X_n$ where each $X_i$ will be one of the $n$ total classes for the attribute i.e the domain categories. 

%We see this means $\bigcup\limits_{i=1}^{n} \text{CC-1-}X_i = \text{CC-1}$. An alternative setting, which we termed \texttt{grouped}, is to disregard the existing data sources and create attribute buckets across the entire corpus, $C$, such that $\bigcup\limits_{i=1}^{n} X_i = C$ as $X_i$ consists of samples among all data sources with that given attribute label. 

\begin{table}[!h]
\centering
  \begin{tabular}{lcc}
   \toprule
      \textbf{Experiment} & \textbf{LM-Eval}  \\
     \midrule
     Baseline & 56.81 \\
     %Baseline 1 (Preference) & 56.81 \\
     %Baseline 2 (UniMax) & 55.83 \\
     \midrule
     Quality \texttt{fine-grained}  & \textbf{\textit{57.88}} \\ 
     Quality \texttt{grouped}  & \textit{57.19}  \\ 
     \midrule
     %Perplexity \texttt{fine-grained}  & \textbf{\textit{57.88}}  \\ 
     %Perplexity \texttt{grouped}  & \textit{57.19}  \\ 
     %Quality \texttt{fine-grained}  & \textit{57.29} \\ 
     %Quality \texttt{grouped}  & \textit{57.13}  \\ 
     %\midrule
     Toxicity \texttt{fine-grained}  & 53.62  \\ 
     Toxicity \texttt{grouped}  & 54.99  \\
     \midrule
     Domain \texttt{fine-grained}  & \textit{57.34}  \\ 
     Domain \texttt{grouped}  & \textit{57.45}  \\
     \midrule
     Type of Speech \texttt{fine-grained}  & 56.69 \\ 
     Type of Speech \texttt{grouped}  & \textit{57.31}  \\
    \bottomrule
  \end{tabular}
  \caption{Sampling weights based on buckets of data attribute labels significantly improve upon baseline results. Italics indicate results that outperform the baseline. Per-task accuracies are shared in Table \ref{tab:full_data_attr_samp}.}
  \label{tab:data_attr_samp}
\end{table}

To assess the utility of attribute based data sampling, we train a 2B model for 165B tokens on a training set of 5 CC snapshots. Our baseline result is when attribute information is not included in data sampling. Further experimental details are shared in Appendix \ref{sec:appendix_data_attribute_experiments}. Table \ref{tab:data_attr_samp} highlights that all attributes aside from toxicity realize improved accuracy when used within data sampling. We note attributes which define broad classes of documents, like domain and type of speech, are more performant in the \texttt{grouped} setting while attributes that assess a characteristic of a document, like quality, are better suited to the \texttt{fine-grained} setting. 

%The sampling weights and full

%Attribute information not compared against each other but to be used to improve, future work to include in the best manner possible

 %When used in sampling we can do the following. The setup consists of blah and blah. Compare against all our classifers ,  As we have info on the CC snapshots we just comapre on those and train 2B models for 165B tokens. Table 1 highlights that comapred to unimax it greatly improve sblah Baselien is unimax vs preference as fine web shows more recent utility to more recent snapshots so as distibrution is fundamentally differnet from earlier study we wnat to benchamrk against best possible baseline. At this token scale and size of CC snapshots unimax will provide nothing more than uniform (note this is markedly differnet setting than compared earlier)

\begin{table}[!h]
\centering
  \begin{tabular}{llc}
   \toprule
      \textbf{Experiment} & \textbf{Target Set} & \textbf{LM-Eval} \\
    \midrule
     Original CC   &  N/A &  54.90 \\
     DSIR  &  Wikipedia, Books &  55.35 \\
     DSIR   &  Low Tox, High Qual &  \textbf{55.63} \\
    \bottomrule
  \end{tabular}
  \caption{Attribute information defines better target sets for data selection. Tox is Toxicity, Qual is Quality.}
  \label{tab:data_attr_sel}
\end{table}

With data attributes, more precise target sets for data selection can be defined. For instance, one with examples that are of both low toxicity and high quality. Table \ref{tab:data_attr_sel} shows that using such a target set with DSIR outperforms the paper-recommended target set and enables toxicity based selection without accuracy degradation.

%Data attributes can also be used to better inform data selection methods on which documents should be removed. We hypothesize that removing toxic examples that also are of low quality would mitigate the past performance degradation seen from toxicity filtering.  Thus, in DSIR we experiment with using a target set of documents which are of low toxicity and high quality. 
%Table \ref{tab:data_attr_sel} shows this target set based on attribute labels is able to outperform the previously best identified setting of DSIR.

Additional angles where data attributes can refine pretraining sets would be through better selection of documents with information amenable for rephrasing \citep{maini2024rephrasing} or seeding synthetic generation pipelines \citep{abdin2024phi3}.

\section{Related Work}

%The data used in training LMs is an integral factor in the resulting quality of the model. 

%In the development process of most recent pretraining sets for language models, data curation, which consists of identification, organization, storage and cleaning of datasets \citep{mclure2014curation, Freitas2016, Thirumuruganathan2020DataCW}, has been the most well-studied and prominent factor. Early models such as BERT \citep{Devlin2019BERTPO},
%and RoBERTa \citep{liu2019roberta} focused on collecting documents from known high quality sources such as Wikipedia while others like XLNet \citep{yang2020xlnet} started to utilize information from Common Crawl web snapshots -- in all cases, the major emphasis in data curation was placed on identification of additional data sources rather than cleaning. To continue realizing improvements in LM capabilities, in conjunction with larger collections of datasets such as C4 \citep{raffel2020exploring}, the Pile \citep{gao2020pile}, and BigScience ROOTS \citep{lachaux2020unsupervised} that began to emerge, heuristic and classifier based filters were used to remove ill-formed and low quality documents \citep{Rae2021Gopher, chowdhery2022palm, raffel2020exploring}. As models that made use of these pretraining sets continued to push the capabilities of existing LMs, further studies within data curation introduced data deduplication \citep{broderdedup, kandpal2022deduplicating, abbas2023semdedup} as a crucial step within the process. 

Data curation, which is the identification, organization, storage and cleaning of datasets \citep{mclure2014curation, Freitas2016, Thirumuruganathan2020DataCW}, has been the most well-studied aspect in pretraining set development. Early models, like BERT \citep{Devlin2019BERTPO} and XLNet \citep{yang2020xlnet}, focused their data curation efforts on obtaining examples from high quality sources. In conjunction with the creation of larger collections of datasets such as C4 \citep{raffel2020exploring}, the Pile \citep{gao2020pile}, and BigScience ROOTS \citep{lachaux2020unsupervised}, heuristic and classifier based filters were used in data curation to remove ill-formed and useless documents \citep{Rae2021Gopher, chowdhery2022palm, raffel2020exploring}. Additional studies within data curation found that data deduplication \citep{broderdedup, kandpal2022deduplicating, abbas2023semdedup} further improved model capabilities by preventing overtraining on a small set of similar examples. 

%Recently, studies such as \citep{penedo2023refinedweb, fine-web} have experimented across all the major axes within data curation to determine the set of filters and deduplication steps that lead to a highly performant pretraining set. 

Data selection and data sampling play major roles in pretraining set construction. Data selection methods \citep{moore-lewis-2010-intelligent, axelrod2017cynical, xie2023data, engstrom2024dsdm} remove low quality documents to retain examples that more closely align with a predetermined high quality source. Moore-Lewis selection \citep{moore-lewis-2010-intelligent} proposed the initial approach, with recent extensions by cynical data selection \citep{axelrod2017cynical} and DSIR \citep{xie2023data} which both better estimate the probability that a given example belongs to a high quality domain. Data sampling techniques either use a learned model \citep{xie2023doremi, albalak2023efficient, fan2024doge} or a heuristic function \citep{arivazhagan2019massively,raffel2020exploring,  chung2023unimax} to define sampling weights for each data source. Learned techniques, such as DoReMi \citep{xie2023doremi}, use the loss of a model across the data sources to define sampling weights while heuristic functions often use the size of a data source to explicitly define weights \citep{arivazhagan2019massively, raffel2020exploring} or  fit a probability distribution \citep{chung2023unimax}.

The data attributes of toxicity and quality have been used to further refine pretraining sets \citep{gururangan2022language, meade2022empirical}. Toxicity classifiers \citep{welbl2021challenges} that remove highly toxic examples reduce the number of toxic generations from LMs, but also negatively impact the model's other capabilities \citep{xu2021detoxifying}. Quality classifiers \citep{Devlin2019BERTPO, raffel2020exploring, chowdhery2022palm} which remove documents such as machine generated texts \citep{dodge2021documenting} or hate speech and sexually explicit content \citep{luccioni2021whats} greatly improve model capabilities. \citep{longpre2023pretrainers} extensively investigate the impact that toxicity, quality, and age of data have on model accuracy.

\section{Conclusion}

We present the first comprehensive study on pretraining set development conducted at the scale of modern day LMs and pretraining set sizes. Through a series of ablations, we identify helpful methods to use at each step of the pretraining development pipeline. We then analyze most currently available web crawl snapshots across the attribute labels of toxicity, quality, domain, and type of speech to better understand the composition of the most widely used data source in current pretraining corpora. These attribute labels are then shown to provide significant improvement in model abilities when incorporated within data selection and data sampling methods. We hope that the open transmission of this knowledge spurns more rapid advancements in the capabilities of LMs. 

%, highlighting that it is more useful to partition pretraining datasets at the granularity of fine-grained attributes rather than at the level of data sources. 

\section*{Limitations}

While we designed our experimental setting to be as generally applicable as possible, we acknowledge that our findings are limited to the distribution of data sources, learning algorithm, and model configuration that we consider. Thus, when extrapolating our findings on pretraining set development to a setting with markedly different data sources or for usage in an alternate type of model, it may be that our results do not hold as strongly. In addition, we do not evaluate all possible techniques for each step of the pretreating pipeline so our results can not be thought of as the definitive rankings amongst all potential methods but rather as a set of strategies with which to create an effective, high-quality pretraining set. Lastly, although the use of synthetic data has recently garnered lots of attention, we did not include any such source of data within our studies and aspects relating to quality selection and sampling of synthetic data may be different than what our findings suggest.

\clearpage
\newpage

\bibliography{custom}

\begin{thebibliography}{100}
\providecommand{\natexlab}[1]{#1}

\bibitem[{Abbas et~al.(2023)Abbas, Tirumala, Simig, Ganguli, and Morcos}]{abbas2023semdedup}
Amro Abbas, Kushal Tirumala, Dániel Simig, Surya Ganguli, and Ari~S. Morcos. 2023.
\newblock \href {https://arxiv.org/abs/2303.09540} {Semdedup: Data-efficient learning at web-scale through semantic deduplication}.
\newblock \emph{Preprint}, arXiv:2303.09540.

\bibitem[{Abdin et~al.(2024)Abdin, Jacobs, Awan, Aneja, Awadallah, Awadalla, Bach, Bahree, Bakhtiari, Behl, Benhaim, Bilenko, Bjorck, Bubeck, Cai, Mendes, Chen, Chaudhary, Chopra, Giorno, de~Rosa, Dixon, Eldan, Iter, Garg, Goswami, Gunasekar, Haider, Hao, Hewett, Huynh, Javaheripi, Jin, Kauffmann, Karampatziakis, Kim, Khademi, Kurilenko, Lee, Lee, Li, Liang, Liu, Lin, Lin, Madan, Mitra, Modi, Nguyen, Norick, Patra, Perez-Becker, Portet, Pryzant, Qin, Radmilac, Rosset, Roy, Ruwase, Saarikivi, Saied, Salim, Santacroce, Shah, Shang, Sharma, Song, Tanaka, Wang, Ward, Wang, Witte, Wyatt, Xu, Xu, Yadav, Yang, Yang, Yu, Zhang, Zhang, Zhang, Zhang, Zhang, Zhang, Zhang, and Zhou}]{abdin2024phi3}
Marah Abdin, Sam~Ade Jacobs, Ammar~Ahmad Awan, Jyoti Aneja, Ahmed Awadallah, Hany Awadalla, Nguyen Bach, Amit Bahree, Arash Bakhtiari, Harkirat Behl, Alon Benhaim, Misha Bilenko, Johan Bjorck, Sébastien Bubeck, Martin Cai, Caio César~Teodoro Mendes, Weizhu Chen, Vishrav Chaudhary, Parul Chopra, Allie~Del Giorno, Gustavo de~Rosa, Matthew Dixon, Ronen Eldan, Dan Iter, Amit Garg, Abhishek Goswami, Suriya Gunasekar, Emman Haider, Junheng Hao, Russell~J. Hewett, Jamie Huynh, Mojan Javaheripi, Xin Jin, Piero Kauffmann, Nikos Karampatziakis, Dongwoo Kim, Mahoud Khademi, Lev Kurilenko, James~R. Lee, Yin~Tat Lee, Yuanzhi Li, Chen Liang, Weishung Liu, Eric Lin, Zeqi Lin, Piyush Madan, Arindam Mitra, Hardik Modi, Anh Nguyen, Brandon Norick, Barun Patra, Daniel Perez-Becker, Thomas Portet, Reid Pryzant, Heyang Qin, Marko Radmilac, Corby Rosset, Sambudha Roy, Olatunji Ruwase, Olli Saarikivi, Amin Saied, Adil Salim, Michael Santacroce, Shital Shah, Ning Shang, Hiteshi Sharma, Xia Song, Masahiro Tanaka, Xin Wang, Rachel
  Ward, Guanhua Wang, Philipp Witte, Michael Wyatt, Can Xu, Jiahang Xu, Sonali Yadav, Fan Yang, Ziyi Yang, Donghan Yu, Chengruidong Zhang, Cyril Zhang, Jianwen Zhang, Li~Lyna Zhang, Yi~Zhang, Yue Zhang, Yunan Zhang, and Xiren Zhou. 2024.
\newblock \href {https://arxiv.org/abs/2404.14219} {Phi-3 technical report: A highly capable language model locally on your phone}.
\newblock \emph{Preprint}, arXiv:2404.14219.

\bibitem[{Albalak et~al.(2024)Albalak, Elazar, Xie, Longpre, Lambert, Wang, Muennighoff, Hou, Pan, Jeong, Raffel, Chang, Hashimoto, and Wang}]{albalak2024survey}
Alon Albalak, Yanai Elazar, Sang~Michael Xie, Shayne Longpre, Nathan Lambert, Xinyi Wang, Niklas Muennighoff, Bairu Hou, Liangming Pan, Haewon Jeong, Colin Raffel, Shiyu Chang, Tatsunori Hashimoto, and William~Yang Wang. 2024.
\newblock \href {https://arxiv.org/abs/2402.16827} {A survey on data selection for language models}.
\newblock \emph{Preprint}, arXiv:2402.16827.

\bibitem[{Albalak et~al.(2023)Albalak, Pan, Raffel, and Wang}]{albalak2023efficient}
Alon Albalak, Liangming Pan, Colin Raffel, and William~Yang Wang. 2023.
\newblock \href {https://arxiv.org/abs/2312.02406} {Efficient online data mixing for language model pre-training}.
\newblock \emph{Preprint}, arXiv:2312.02406.

\bibitem[{Allal et~al.(2023)Allal, Li, Kocetkov, Mou, Akiki, Ferrandis, Muennighoff, Mishra, Gu, Dey, Umapathi, Anderson, Zi, Poirier, Schoelkopf, Troshin, Abulkhanov, Romero, Lappert, Toni, del Río, Liu, Bose, Bhattacharyya, Zhuo, Yu, Villegas, Zocca, Mangrulkar, Lansky, Nguyen, Contractor, Villa, Li, Bahdanau, Jernite, Hughes, Fried, Guha, de~Vries, and von Werra}]{allal2023santacoder}
Loubna~Ben Allal, Raymond Li, Denis Kocetkov, Chenghao Mou, Christopher Akiki, Carlos~Munoz Ferrandis, Niklas Muennighoff, Mayank Mishra, Alex Gu, Manan Dey, Logesh~Kumar Umapathi, Carolyn~Jane Anderson, Yangtian Zi, Joel~Lamy Poirier, Hailey Schoelkopf, Sergey Troshin, Dmitry Abulkhanov, Manuel Romero, Michael Lappert, Francesco~De Toni, Bernardo~García del Río, Qian Liu, Shamik Bose, Urvashi Bhattacharyya, Terry~Yue Zhuo, Ian Yu, Paulo Villegas, Marco Zocca, Sourab Mangrulkar, David Lansky, Huu Nguyen, Danish Contractor, Luis Villa, Jia Li, Dzmitry Bahdanau, Yacine Jernite, Sean Hughes, Daniel Fried, Arjun Guha, Harm de~Vries, and Leandro von Werra. 2023.
\newblock \href {https://arxiv.org/abs/2301.03988} {Santacoder: don't reach for the stars!}
\newblock \emph{Preprint}, arXiv:2301.03988.

\bibitem[{Almazrouei et~al.(2023)Almazrouei, Alobeidli, Alshamsi, Cappelli, Cojocaru, Debbah, Étienne Goffinet, Hesslow, Launay, Malartic, Mazzotta, Noune, Pannier, and Penedo}]{almazrouei2023falcon}
Ebtesam Almazrouei, Hamza Alobeidli, Abdulaziz Alshamsi, Alessandro Cappelli, Ruxandra Cojocaru, Mérouane Debbah, Étienne Goffinet, Daniel Hesslow, Julien Launay, Quentin Malartic, Daniele Mazzotta, Badreddine Noune, Baptiste Pannier, and Guilherme Penedo. 2023.
\newblock \href {https://arxiv.org/abs/2311.16867} {The falcon series of open language models}.
\newblock \emph{Preprint}, arXiv:2311.16867.

\bibitem[{Anthropic(2024)}]{claude32024}
Anthropic. 2024.
\newblock {The Claude 3 Model Family: Opus, Sonnet, Haiku}.

\bibitem[{Arivazhagan et~al.(2019)Arivazhagan, Bapna, Firat, Lepikhin, Johnson, Krikun, Chen, Cao, Foster, Cherry, Macherey, Chen, and Wu}]{arivazhagan2019massively}
Naveen Arivazhagan, Ankur Bapna, Orhan Firat, Dmitry Lepikhin, Melvin Johnson, Maxim Krikun, Mia~Xu Chen, Yuan Cao, George Foster, Colin Cherry, Wolfgang Macherey, Zhifeng Chen, and Yonghui Wu. 2019.
\newblock \href {https://arxiv.org/abs/1907.05019} {Massively multilingual neural machine translation in the wild: Findings and challenges}.
\newblock \emph{Preprint}, arXiv:1907.05019.

\bibitem[{Axelrod(2017)}]{axelrod2017cynical}
Amittai Axelrod. 2017.
\newblock \href {https://arxiv.org/abs/1709.02279} {Cynical selection of language model training data}.
\newblock \emph{Preprint}, arXiv:1709.02279.

\bibitem[{Bai et~al.(2023)Bai, Bai, Chu, Cui, Dang, Deng, Fan, Ge, Han, Huang et~al.}]{bai2023qwen}
Jinze Bai, Shuai Bai, Yunfei Chu, Zeyu Cui, Kai Dang, Xiaodong Deng, Yang Fan, Wenbin Ge, Yu~Han, Fei Huang, et~al. 2023.
\newblock {Qwen Technical Report}.
\newblock \emph{arXiv preprint arXiv:2309.16609}.

\bibitem[{Baumgartner et~al.(2020)Baumgartner, Zannettou, Keegan, Squire, and Blackburn}]{reddit-dataset-2020}
Jason Baumgartner, Savvas Zannettou, Brian Keegan, Megan Squire, and Jeremy Blackburn. 2020.
\newblock The pushshift reddit dataset.
\newblock In \emph{Proceedings of the international AAAI conference on web and social media}, volume~14, pages 830--839.

\bibitem[{BigScience(2023)}]{workshop2023bloom}
BigScience. 2023.
\newblock \href {https://arxiv.org/abs/2211.05100} {{BLOOM: A 176B-Parameter Open-Access Multilingual Language Model}}.
\newblock \emph{Preprint}, arXiv:2211.05100.

\bibitem[{Bisk et~al.(2020)Bisk, Zellers, Bras, Gao, and Choi}]{Bisk2020PIQARA}
Yonatan Bisk, Rowan Zellers, Ronan~Le Bras, Jianfeng Gao, and Yejin Choi. 2020.
\newblock Piqa: Reasoning about physical commonsense in natural language.
\newblock In \emph{AAAI}.

\bibitem[{Broder(1997)}]{broderdedup}
A.Z. Broder. 1997.
\newblock \href {https://doi.org/10.1109/SEQUEN.1997.666900} {On the resemblance and containment of documents}.
\newblock In \emph{Proceedings. Compression and Complexity of SEQUENCES 1997 (Cat. No.97TB100171)}, pages 21--29.

\bibitem[{Brown et~al.(2020)Brown, Mann, Ryder, Subbiah, Kaplan, Dhariwal, Neelakantan, Shyam, Sastry, Askell, Agarwal, Herbert{-}Voss, Krueger, Henighan, Child, Ramesh, Ziegler, Wu, Winter, Hesse, Chen, Sigler, Litwin, Gray, Chess, Clark, Berner, McCandlish, Radford, Sutskever, and Amodei}]{brown2020language}
Tom~B. Brown, Benjamin Mann, Nick Ryder, Melanie Subbiah, Jared Kaplan, Prafulla Dhariwal, Arvind Neelakantan, Pranav Shyam, Girish Sastry, Amanda Askell, Sandhini Agarwal, Ariel Herbert{-}Voss, Gretchen Krueger, Tom Henighan, Rewon Child, Aditya Ramesh, Daniel~M. Ziegler, Jeffrey Wu, Clemens Winter, Christopher Hesse, Mark Chen, Eric Sigler, Mateusz Litwin, Scott Gray, Benjamin Chess, Jack Clark, Christopher Berner, Sam McCandlish, Alec Radford, Ilya Sutskever, and Dario Amodei. 2020.
\newblock \href {https://proceedings.neurips.cc/paper/2020/hash/1457c0d6bfcb4967418bfb8ac142f64a-Abstract.html} {Language models are few-shot learners}.
\newblock In \emph{Advances in Neural Information Processing Systems 33: Annual Conference on Neural Information Processing Systems 2020, NeurIPS 2020, December 6-12, 2020, virtual}.

\bibitem[{Cassano et~al.(2023)Cassano, Gouwar, Nguyen, Nguyen, Phipps-Costin, Pinckney, Yee, Zi, Anderson, Feldman, Guha, Greenberg, and Jangda}]{multiple2023}
Federico Cassano, John Gouwar, Daniel Nguyen, Sydney Nguyen, Luna Phipps-Costin, Donald Pinckney, Ming-Ho Yee, Yangtian Zi, Carolyn~Jane Anderson, Molly~Q Feldman, Arjun Guha, Michael Greenberg, and Abhinav Jangda. 2023.
\newblock \href {https://doi.org/10.1109/TSE.2023.3267446} {Multipl-e: A scalable and polyglot approach to benchmarking neural code generation}.
\newblock \emph{IEEE Transactions on Software Engineering}, pages 1--17.

\bibitem[{Chawla et~al.(2021)Chawla, Ramirez, Clever, Lucas, May, and Gratch}]{chawla2021casino}
Kushal Chawla, Jaysa Ramirez, Rene Clever, Gale Lucas, Jonathan May, and Jonathan Gratch. 2021.
\newblock Casino: A corpus of campsite negotiation dialogues for automatic negotiation systems.
\newblock \emph{arXiv preprint arXiv:2103.15721}.

\bibitem[{Chen et~al.(2021)Chen, Tworek, Jun, Yuan, de~Oliveira~Pinto, Kaplan, Edwards, Burda, Joseph, Brockman, Ray, Puri, Krueger, Petrov, Khlaaf, Sastry, Mishkin, Chan, Gray, Ryder, Pavlov, Power, Kaiser, Bavarian, Winter, Tillet, Such, Cummings, Plappert, Chantzis, Barnes, Herbert-Voss, Guss, Nichol, Paino, Tezak, Tang, Babuschkin, Balaji, Jain, Saunders, Hesse, Carr, Leike, Achiam, Misra, Morikawa, Radford, Knight, Brundage, Murati, Mayer, Welinder, McGrew, Amodei, McCandlish, Sutskever, and Zaremba}]{chen2021evaluating}
Mark Chen, Jerry Tworek, Heewoo Jun, Qiming Yuan, Henrique~Ponde de~Oliveira~Pinto, Jared Kaplan, Harri Edwards, Yuri Burda, Nicholas Joseph, Greg Brockman, Alex Ray, Raul Puri, Gretchen Krueger, Michael Petrov, Heidy Khlaaf, Girish Sastry, Pamela Mishkin, Brooke Chan, Scott Gray, Nick Ryder, Mikhail Pavlov, Alethea Power, Lukasz Kaiser, Mohammad Bavarian, Clemens Winter, Philippe Tillet, Felipe~Petroski Such, Dave Cummings, Matthias Plappert, Fotios Chantzis, Elizabeth Barnes, Ariel Herbert-Voss, William~Hebgen Guss, Alex Nichol, Alex Paino, Nikolas Tezak, Jie Tang, Igor Babuschkin, Suchir Balaji, Shantanu Jain, William Saunders, Christopher Hesse, Andrew~N. Carr, Jan Leike, Josh Achiam, Vedant Misra, Evan Morikawa, Alec Radford, Matthew Knight, Miles Brundage, Mira Murati, Katie Mayer, Peter Welinder, Bob McGrew, Dario Amodei, Sam McCandlish, Ilya Sutskever, and Wojciech Zaremba. 2021.
\newblock \href {https://arxiv.org/abs/2107.03374} {Evaluating large language models trained on code}.
\newblock \emph{Preprint}, arXiv:2107.03374.

\bibitem[{Chowdhery et~al.(2022)Chowdhery, Narang, Devlin, Bosma, Mishra, Roberts, Barham, Chung, Sutton, Gehrmann et~al.}]{chowdhery2022palm}
Aakanksha Chowdhery, Sharan Narang, Jacob Devlin, Maarten Bosma, Gaurav Mishra, Adam Roberts, Paul Barham, Hyung~Won Chung, Charles Sutton, Sebastian Gehrmann, et~al. 2022.
\newblock {PaLM: Scaling Language Modeling with Pathways}.
\newblock \emph{arXiv preprint arXiv:2204.02311}.

\bibitem[{Chung et~al.(2023)Chung, Constant, Garcia, Roberts, Tay, Narang, and Firat}]{chung2023unimax}
Hyung~Won Chung, Noah Constant, Xavier Garcia, Adam Roberts, Yi~Tay, Sharan Narang, and Orhan Firat. 2023.
\newblock \href {https://arxiv.org/abs/2304.09151} {Unimax: Fairer and more effective language sampling for large-scale multilingual pretraining}.
\newblock \emph{Preprint}, arXiv:2304.09151.

\bibitem[{Clark et~al.(2020)Clark, Choi, Collins, Garrette, Kwiatkowski, Nikolaev, and Palomaki}]{tydiqa}
Jonathan~H. Clark, Eunsol Choi, Michael Collins, Dan Garrette, Tom Kwiatkowski, Vitaly Nikolaev, and Jennimaria Palomaki. 2020.
\newblock \href {https://arxiv.org/abs/2003.05002} {Tydi {QA:} {A} benchmark for information-seeking question answering in typologically diverse languages}.
\newblock \emph{CoRR}, abs/2003.05002.

\bibitem[{Clark et~al.(2018)Clark, Cowhey, Etzioni, Khot, Sabharwal, Schoenick, and Tafjord}]{clark2018think}
Peter Clark, Isaac Cowhey, Oren Etzioni, Tushar Khot, Ashish Sabharwal, Carissa Schoenick, and Oyvind Tafjord. 2018.
\newblock Think you have solved question answering? try arc, the ai2 reasoning challenge.
\newblock \emph{arXiv preprint arXiv:1803.05457}.

\bibitem[{Devlin et~al.(2019)Devlin, Chang, Lee, and Toutanova}]{Devlin2019BERTPO}
Jacob Devlin, Ming-Wei Chang, Kenton Lee, and Kristina Toutanova. 2019.
\newblock Bert: Pre-training of deep bidirectional transformers for language understanding.
\newblock In \emph{NAACL}.

\bibitem[{Dodge et~al.(2021)Dodge, Sap, Marasović, Agnew, Ilharco, Groeneveld, Mitchell, and Gardner}]{dodge2021documenting}
Jesse Dodge, Maarten Sap, Ana Marasović, William Agnew, Gabriel Ilharco, Dirk Groeneveld, Margaret Mitchell, and Matt Gardner. 2021.
\newblock \href {https://arxiv.org/abs/2104.08758} {Documenting large webtext corpora: A case study on the colossal clean crawled corpus}.
\newblock \emph{Preprint}, arXiv:2104.08758.

\bibitem[{El-Kishky et~al.(2019)El-Kishky, Chaudhary, Guzm{\'a}n, and Koehn}]{el2019ccaligned}
Ahmed El-Kishky, Vishrav Chaudhary, Francisco Guzm{\'a}n, and Philipp Koehn. 2019.
\newblock Ccaligned: A massive collection of cross-lingual web-document pairs.
\newblock \emph{arXiv preprint arXiv:1911.06154}.

\bibitem[{Engstrom et~al.(2024)Engstrom, Feldmann, and Madry}]{engstrom2024dsdm}
Logan Engstrom, Axel Feldmann, and Aleksander Madry. 2024.
\newblock \href {https://arxiv.org/abs/2401.12926} {Dsdm: Model-aware dataset selection with datamodels}.
\newblock \emph{Preprint}, arXiv:2401.12926.

\bibitem[{Espl{\`a}-Gomis et~al.(2019)Espl{\`a}-Gomis, Forcada, Ram{\'\i}rez-S{\'a}nchez, and Hoang}]{espla2019paracrawl}
Miquel Espl{\`a}-Gomis, Mikel~L Forcada, Gema Ram{\'\i}rez-S{\'a}nchez, and Hieu Hoang. 2019.
\newblock Paracrawl: Web-scale parallel corpora for the languages of the eu.
\newblock In \emph{Proceedings of Machine Translation Summit XVII: Translator, Project and User Tracks}, pages 118--119.

\bibitem[{Fan et~al.(2024)Fan, Pagliardini, and Jaggi}]{fan2024doge}
Simin Fan, Matteo Pagliardini, and Martin Jaggi. 2024.
\newblock \href {https://arxiv.org/abs/2310.15393} {Doge: Domain reweighting with generalization estimation}.
\newblock \emph{Preprint}, arXiv:2310.15393.

\bibitem[{Ferschke et~al.(2012)Ferschke, Gurevych, and Chebotar}]{ferschke2012behind}
Oliver Ferschke, Iryna Gurevych, and Yevgen Chebotar. 2012.
\newblock Behind the article: Recognizing dialog acts in wikipedia talk pages.
\newblock In \emph{Proceedings of the 13th Conference of the European Chapter of the Association for Computational Linguistics}, pages 777--786.

\bibitem[{Freitas and Curry(2016)}]{Freitas2016}
Andr{\'e} Freitas and Edward Curry. 2016.
\newblock \href {https://doi.org/10.1007/978-3-319-21569-3_6} {\emph{Big Data Curation}}, pages 87--118.
\newblock Springer International Publishing, Cham.

\bibitem[{Gao(2021)}]{gao2021empirical}
Leo Gao. 2021.
\newblock An empirical exploration in quality filtering of text data.
\newblock \emph{arXiv preprint arXiv:2109.00698}.

\bibitem[{Gao et~al.(2020{\natexlab{a}})Gao, Biderman, Black, Golding, Hoppe, Foster, Phang, He, Thite, Nabeshima, Presser, and Leahy}]{gao2020pile}
Leo Gao, Stella Biderman, Sid Black, Laurence Golding, Travis Hoppe, Charles Foster, Jason Phang, Horace He, Anish Thite, Noa Nabeshima, Shawn Presser, and Connor Leahy. 2020{\natexlab{a}}.
\newblock \href {https://arxiv.org/abs/2101.00027} {The pile: An 800gb dataset of diverse text for language modeling}.
\newblock \emph{Preprint}, arXiv:2101.00027.

\bibitem[{Gao et~al.(2020{\natexlab{b}})Gao, Biderman, Black, Golding, Hoppe, Foster, Phang, He, Thite, Nabeshima, Presser, and Leahy}]{pile-dataset-2020}
Leo Gao, Stella Biderman, Sid Black, Laurence Golding, Travis Hoppe, Charles Foster, Jason Phang, Horace He, Anish Thite, Noa Nabeshima, Shawn Presser, and Connor Leahy. 2020{\natexlab{b}}.
\newblock The {P}ile: An 800gb dataset of diverse text for language modeling.
\newblock \emph{arXiv preprint arXiv:2101.00027}.

\bibitem[{Gemma~Team(2024)}]{gemma24}
Google~DeepMind Gemma~Team. 2024.
\newblock {Gemma: Open Models Based on Gemini Research and Technology}.

\bibitem[{Gururangan et~al.(2022)Gururangan, Card, Dreier, Gade, Wang, Wang, Zettlemoyer, and Smith}]{gururangan2022language}
Suchin Gururangan, Dallas Card, Sarah~K. Dreier, Emily~K. Gade, Leroy~Z. Wang, Zeyu Wang, Luke Zettlemoyer, and Noah~A. Smith. 2022.
\newblock \href {https://arxiv.org/abs/2201.10474} {Whose language counts as high quality? measuring language ideologies in text data selection}.
\newblock \emph{Preprint}, arXiv:2201.10474.

\bibitem[{Heafield(2011)}]{heafield-2011-kenlm}
Kenneth Heafield. 2011.
\newblock \href {https://aclanthology.org/W11-2123} {{K}en{LM}: Faster and smaller language model queries}.
\newblock In \emph{Proceedings of the Sixth Workshop on Statistical Machine Translation}, pages 187--197, Edinburgh, Scotland. Association for Computational Linguistics.

\bibitem[{Henderson et~al.(2022)Henderson, Krass, Zheng, Guha, Manning, Jurafsky, and Ho}]{henderson2022pile}
Peter Henderson, Mark~S. Krass, Lucia Zheng, Neel Guha, Christopher~D. Manning, Dan Jurafsky, and Daniel~E. Ho. 2022.
\newblock \href {https://arxiv.org/abs/2207.00220} {Pile of law: Learning responsible data filtering from the law and a 256gb open-source legal dataset}.
\newblock \emph{Preprint}, arXiv:2207.00220.

\bibitem[{Hendrycks et~al.(2020)Hendrycks, Burns, Basart, Zou, Mazeika, Song, and Steinhardt}]{hendrycks2020measuring}
Dan Hendrycks, Collin Burns, Steven Basart, Andy Zou, Mantas Mazeika, Dawn Song, and Jacob Steinhardt. 2020.
\newblock Measuring massive multitask language understanding.
\newblock \emph{arXiv preprint arXiv:2009.03300}.

\bibitem[{Hoffmann et~al.(2022)Hoffmann, Borgeaud, Mensch, Buchatskaya, Cai, Rutherford, Casas, Hendricks, Welbl, Clark et~al.}]{hoffmann2022training}
Jordan Hoffmann, Sebastian Borgeaud, Arthur Mensch, Elena Buchatskaya, Trevor Cai, Eliza Rutherford, Diego de~Las Casas, Lisa~Anne Hendricks, Johannes Welbl, Aidan Clark, et~al. 2022.
\newblock {Training Compute-Optimal Large Language Models}.
\newblock \emph{arXiv preprint arXiv:2203.15556}.

\bibitem[{Hu et~al.(2020)Hu, Ruder, Siddhant, Neubig, Firat, and Johnson}]{hu2020xtreme}
Junjie Hu, Sebastian Ruder, Aditya Siddhant, Graham Neubig, Orhan Firat, and Melvin Johnson. 2020.
\newblock \href {https://arxiv.org/abs/2003.11080} {Xtreme: A massively multilingual multi-task benchmark for evaluating cross-lingual generalization}.
\newblock \emph{Preprint}, arXiv:2003.11080.

\bibitem[{Jiang et~al.(2023)Jiang, Sablayrolles, Mensch, Bamford, Chaplot, Casas, Bressand, Lengyel, Lample, Saulnier et~al.}]{jiang2023mistral}
Albert~Q Jiang, Alexandre Sablayrolles, Arthur Mensch, Chris Bamford, Devendra~Singh Chaplot, Diego de~las Casas, Florian Bressand, Gianna Lengyel, Guillaume Lample, Lucile Saulnier, et~al. 2023.
\newblock {Mistral 7B}.
\newblock \emph{arXiv preprint arXiv:2310.06825}.

\bibitem[{K et~al.(2020)K, Wang, Mayhew, and Roth}]{k2020crosslingual}
Karthikeyan K, Zihan Wang, Stephen Mayhew, and Dan Roth. 2020.
\newblock \href {https://arxiv.org/abs/1912.07840} {Cross-lingual ability of multilingual bert: An empirical study}.
\newblock \emph{Preprint}, arXiv:1912.07840.

\bibitem[{Kandpal et~al.(2022)Kandpal, Wallace, and Raffel}]{kandpal2022deduplicating}
Nikhil Kandpal, Eric Wallace, and Colin Raffel. 2022.
\newblock \href {https://arxiv.org/abs/2202.06539} {Deduplicating training data mitigates privacy risks in language models}.
\newblock \emph{Preprint}, arXiv:2202.06539.

\bibitem[{Kocetkov et~al.(2022)Kocetkov, Li, Allal, Li, Mou, Ferrandis, Jernite, Mitchell, Hughes, Wolf et~al.}]{kocetkov2022stack}
Denis Kocetkov, Raymond Li, Loubna~Ben Allal, Jia Li, Chenghao Mou, Carlos~Mu{\~n}oz Ferrandis, Yacine Jernite, Margaret Mitchell, Sean Hughes, Thomas Wolf, et~al. 2022.
\newblock The stack: 3 tb of permissively licensed source code.
\newblock \emph{arXiv preprint arXiv:2211.15533}.

\bibitem[{Kudo and Richardson(2018)}]{kudo2018sentencepiece}
Taku Kudo and John Richardson. 2018.
\newblock Sentencepiece: A simple and language independent subword tokenizer and detokenizer for neural text processing.
\newblock \emph{arXiv preprint arXiv:1808.06226}.

\bibitem[{Lachaux et~al.(2020)Lachaux, Roziere, Chanussot, and Lample}]{lachaux2020unsupervised}
Marie-Anne Lachaux, Baptiste Roziere, Lowik Chanussot, and Guillaume Lample. 2020.
\newblock \href {https://arxiv.org/abs/2006.03511} {Unsupervised translation of programming languages}.
\newblock \emph{Preprint}, arXiv:2006.03511.

\bibitem[{Lai et~al.(2017)Lai, Xie, Liu, Yang, and Hovy}]{lai-etal-2017-race}
Guokun Lai, Qizhe Xie, Hanxiao Liu, Yiming Yang, and Eduard Hovy. 2017.
\newblock \href {https://doi.org/10.18653/v1/D17-1082} {{RACE}: Large-scale {R}e{A}ding comprehension dataset from examinations}.
\newblock In \emph{Proceedings of the 2017 Conference on Empirical Methods in Natural Language Processing}, pages 785--794, Copenhagen, Denmark. Association for Computational Linguistics.

\bibitem[{Lauren\c{c}on et~al.(2022)Lauren\c{c}on, Saulnier, Wang, Akiki, Villanova~del Moral, Le~Scao, Von~Werra, Mou, Gonz\'{a}lez~Ponferrada, Nguyen, Frohberg, \v{S}a\v{s}ko, Lhoest, McMillan-Major, Dupont, Biderman, Rogers, Ben~allal, De~Toni, Pistilli, Nguyen, Nikpoor, Masoud, Colombo, de~la Rosa, Villegas, Thrush, Longpre, Nagel, Weber, Mu\~{n}oz, Zhu, Van~Strien, Alyafeai, Almubarak, Vu, Gonzalez-Dios, Soroa, Lo, Dey, Ortiz~Suarez, Gokaslan, Bose, Adelani, Phan, Tran, Yu, Pai, Chim, Lepercq, Ilic, Mitchell, Luccioni, and Jernite}]{bigscience-dataset-2022}
Hugo Lauren\c{c}on, Lucile Saulnier, Thomas Wang, Christopher Akiki, Albert Villanova~del Moral, Teven Le~Scao, Leandro Von~Werra, Chenghao Mou, Eduardo Gonz\'{a}lez~Ponferrada, Huu Nguyen, J\"{o}rg Frohberg, Mario \v{S}a\v{s}ko, Quentin Lhoest, Angelina McMillan-Major, Gerard Dupont, Stella Biderman, Anna Rogers, Loubna Ben~allal, Francesco De~Toni, Giada Pistilli, Olivier Nguyen, Somaieh Nikpoor, Maraim Masoud, Pierre Colombo, Javier de~la Rosa, Paulo Villegas, Tristan Thrush, Shayne Longpre, Sebastian Nagel, Leon Weber, Manuel Mu\~{n}oz, Jian Zhu, Daniel Van~Strien, Zaid Alyafeai, Khalid Almubarak, Minh~Chien Vu, Itziar Gonzalez-Dios, Aitor Soroa, Kyle Lo, Manan Dey, Pedro Ortiz~Suarez, Aaron Gokaslan, Shamik Bose, David Adelani, Long Phan, Hieu Tran, Ian Yu, Suhas Pai, Jenny Chim, Violette Lepercq, Suzana Ilic, Margaret Mitchell, Sasha~Alexandra Luccioni, and Yacine Jernite. 2022.
\newblock \href {https://proceedings.neurips.cc/paper_files/paper/2022/file/ce9e92e3de2372a4b93353eb7f3dc0bd-Paper-Datasets_and_Benchmarks.pdf} {The bigscience roots corpus: A 1.6tb composite multilingual dataset}.
\newblock In \emph{Advances in Neural Information Processing Systems}, volume~35, pages 31809--31826. Curran Associates, Inc.

\bibitem[{Li et~al.(2023)Li, Allal, Zi, Muennighoff, Kocetkov, Mou, Marone, Akiki, Li, Chim, Liu, Zheltonozhskii, Zhuo, Wang, Dehaene, Davaadorj, Lamy-Poirier, Monteiro, Shliazhko, Gontier, Meade, Zebaze, Yee, Umapathi, Zhu, Lipkin, Oblokulov, Wang, Murthy, Stillerman, Patel, Abulkhanov, Zocca, Dey, Zhang, Fahmy, Bhattacharyya, Yu, Singh, Luccioni, Villegas, Kunakov, Zhdanov, Romero, Lee, Timor, Ding, Schlesinger, Schoelkopf, Ebert, Dao, Mishra, Gu, Robinson, Anderson, Dolan-Gavitt, Contractor, Reddy, Fried, Bahdanau, Jernite, Ferrandis, Hughes, Wolf, Guha, von Werra, and de~Vries}]{li2023starcoder}
Raymond Li, Loubna~Ben Allal, Yangtian Zi, Niklas Muennighoff, Denis Kocetkov, Chenghao Mou, Marc Marone, Christopher Akiki, Jia Li, Jenny Chim, Qian Liu, Evgenii Zheltonozhskii, Terry~Yue Zhuo, Thomas Wang, Olivier Dehaene, Mishig Davaadorj, Joel Lamy-Poirier, João Monteiro, Oleh Shliazhko, Nicolas Gontier, Nicholas Meade, Armel Zebaze, Ming-Ho Yee, Logesh~Kumar Umapathi, Jian Zhu, Benjamin Lipkin, Muhtasham Oblokulov, Zhiruo Wang, Rudra Murthy, Jason Stillerman, Siva~Sankalp Patel, Dmitry Abulkhanov, Marco Zocca, Manan Dey, Zhihan Zhang, Nour Fahmy, Urvashi Bhattacharyya, Wenhao Yu, Swayam Singh, Sasha Luccioni, Paulo Villegas, Maxim Kunakov, Fedor Zhdanov, Manuel Romero, Tony Lee, Nadav Timor, Jennifer Ding, Claire Schlesinger, Hailey Schoelkopf, Jan Ebert, Tri Dao, Mayank Mishra, Alex Gu, Jennifer Robinson, Carolyn~Jane Anderson, Brendan Dolan-Gavitt, Danish Contractor, Siva Reddy, Daniel Fried, Dzmitry Bahdanau, Yacine Jernite, Carlos~Muñoz Ferrandis, Sean Hughes, Thomas Wolf, Arjun Guha, Leandro von
  Werra, and Harm de~Vries. 2023.
\newblock \href {https://arxiv.org/abs/2305.06161} {Starcoder: may the source be with you!}
\newblock \emph{Preprint}, arXiv:2305.06161.

\bibitem[{Liu et~al.(2019)Liu, Ott, Goyal, Du, Joshi, Chen, Levy, Lewis, Zettlemoyer, and Stoyanov}]{liu2019roberta}
Yinhan Liu, Myle Ott, Naman Goyal, Jingfei Du, Mandar Joshi, Danqi Chen, Omer Levy, Mike Lewis, Luke Zettlemoyer, and Veselin Stoyanov. 2019.
\newblock Roberta: A robustly optimized bert pretraining approach.
\newblock \emph{arXiv preprint arXiv:1907.11692}.

\bibitem[{Longpre et~al.(2023)Longpre, Yauney, Reif, Lee, Roberts, Zoph, Zhou, Wei, Robinson, Mimno, and Ippolito}]{longpre2023pretrainers}
Shayne Longpre, Gregory Yauney, Emily Reif, Katherine Lee, Adam Roberts, Barret Zoph, Denny Zhou, Jason Wei, Kevin Robinson, David Mimno, and Daphne Ippolito. 2023.
\newblock \href {https://arxiv.org/abs/2305.13169} {A pretrainer's guide to training data: Measuring the effects of data age, domain coverage, quality, \& toxicity}.
\newblock \emph{Preprint}, arXiv:2305.13169.

\bibitem[{Loshchilov and Hutter(2019)}]{loshchilov2019decoupled}
Ilya Loshchilov and Frank Hutter. 2019.
\newblock \href {https://arxiv.org/abs/1711.05101} {Decoupled weight decay regularization}.
\newblock \emph{Preprint}, arXiv:1711.05101.

\bibitem[{Lowe et~al.(2015)Lowe, Pow, Serban, and Pineau}]{lowe2015ubuntu}
Ryan Lowe, Nissan Pow, Iulian Serban, and Joelle Pineau. 2015.
\newblock The ubuntu dialogue corpus: A large dataset for research in unstructured multi-turn dialogue systems.
\newblock \emph{arXiv preprint arXiv:1506.08909}.

\bibitem[{Luccioni and Viviano(2021)}]{luccioni2021whats}
Alexandra~Sasha Luccioni and Joseph~D. Viviano. 2021.
\newblock \href {https://arxiv.org/abs/2105.02732} {What's in the box? a preliminary analysis of undesirable content in the common crawl corpus}.
\newblock \emph{Preprint}, arXiv:2105.02732.

\bibitem[{Maini et~al.(2024)Maini, Seto, Bai, Grangier, Zhang, and Jaitly}]{maini2024rephrasing}
Pratyush Maini, Skyler Seto, He~Bai, David Grangier, Yizhe Zhang, and Navdeep Jaitly. 2024.
\newblock \href {https://arxiv.org/abs/2401.16380} {Rephrasing the web: A recipe for compute and data-efficient language modeling}.
\newblock \emph{Preprint}, arXiv:2401.16380.

\bibitem[{McLure et~al.(2014)McLure, Level, Cranston, Oehlerts, and Culbertson}]{mclure2014curation}
Merinda McLure, Allison Level, Catherine Cranston, Beth Oehlerts, and Mike Culbertson. 2014.
\newblock \href {https://doi.org/10.1353/pla.2014.0009} {Data curation: A study of researcher practices and needs}.
\newblock \emph{portal: Libraries and the Academy}, 14:139--164.

\bibitem[{Meade et~al.(2022)Meade, Poole-Dayan, and Reddy}]{meade2022empirical}
Nicholas Meade, Elinor Poole-Dayan, and Siva Reddy. 2022.
\newblock \href {https://arxiv.org/abs/2110.08527} {An empirical survey of the effectiveness of debiasing techniques for pre-trained language models}.
\newblock \emph{Preprint}, arXiv:2110.08527.

\bibitem[{Meyers et~al.(2018)Meyers, Munaiah, Prud’hommeaux, Meneely, Wolff, Alm, and Murukannaiah}]{meyers2018dataset}
Benjamin~S Meyers, Nuthan Munaiah, Emily Prud’hommeaux, Andrew Meneely, Josephine Wolff, Cecilia~Ovesdotter Alm, and Pradeep Murukannaiah. 2018.
\newblock A dataset for identifying actionable feedback in collaborative software development.
\newblock In \emph{Proceedings of the 56th Annual Meeting of the Association for Computational Linguistics (Volume 2: Short Papers)}, pages 126--131.

\bibitem[{Moore and Lewis(2010)}]{moore-lewis-2010-intelligent}
Robert~C. Moore and William Lewis. 2010.
\newblock \href {https://aclanthology.org/P10-2041} {Intelligent selection of language model training data}.
\newblock In \emph{Proceedings of the {ACL} 2010 Conference Short Papers}, pages 220--224, Uppsala, Sweden. Association for Computational Linguistics.

\bibitem[{OpenAI(2024)}]{openai2024gpt4}
OpenAI. 2024.
\newblock \href {https://arxiv.org/abs/2303.08774} {Gpt-4 technical report}.
\newblock \emph{Preprint}, arXiv:2303.08774.

\bibitem[{Paperno et~al.(2016)Paperno, Kruszewski, Lazaridou, Pham, Bernardi, Pezzelle, Baroni, Boleda, and Fern{\'a}ndez}]{paperno-etal-2016-lambada}
Denis Paperno, Germ{\'a}n Kruszewski, Angeliki Lazaridou, Ngoc~Quan Pham, Raffaella Bernardi, Sandro Pezzelle, Marco Baroni, Gemma Boleda, and Raquel Fern{\'a}ndez. 2016.
\newblock \href {https://doi.org/10.18653/v1/P16-1144} {The {LAMBADA} dataset: Word prediction requiring a broad discourse context}.
\newblock In \emph{Proceedings of the 54th Annual Meeting of the Association for Computational Linguistics (Volume 1: Long Papers)}, pages 1525--1534, Berlin, Germany. Association for Computational Linguistics.

\bibitem[{Parmar et~al.(2024)Parmar, Prabhumoye, Jennings, Patwary, Subramanian, Su, Zhu, Narayanan, Jhunjhunwala, Dattagupta, Jawa, Liu, Mahabaleshwarkar, Nitski, Brundyn, Maki, Martinez, You, Kamalu, LeGresley, Fridman, Casper, Aithal, Kuchaiev, Shoeybi, Cohen, and Catanzaro}]{parmar2024nemotron4}
Jupinder Parmar, Shrimai Prabhumoye, Joseph Jennings, Mostofa Patwary, Sandeep Subramanian, Dan Su, Chen Zhu, Deepak Narayanan, Aastha Jhunjhunwala, Ayush Dattagupta, Vibhu Jawa, Jiwei Liu, Ameya Mahabaleshwarkar, Osvald Nitski, Annika Brundyn, James Maki, Miguel Martinez, Jiaxuan You, John Kamalu, Patrick LeGresley, Denys Fridman, Jared Casper, Ashwath Aithal, Oleksii Kuchaiev, Mohammad Shoeybi, Jonathan Cohen, and Bryan Catanzaro. 2024.
\newblock \href {https://arxiv.org/abs/2402.16819} {Nemotron-4 15b technical report}.
\newblock \emph{Preprint}, arXiv:2402.16819.

\bibitem[{Penedo et~al.(2024)Penedo, Kydlíček, Ben~Allal, Lozhkov, Raffel, Werra, and Wolf}]{fine-web}
Guilherme Penedo, Hynek Kydlíček, Loubna Ben~Allal, Anton Lozhkov, Colin Raffel, Leandro Werra, and Thomas Wolf. 2024.
\newblock \href {https://huggingface.co/spaces/HuggingFaceFW/blogpost-fineweb-v1} {{FineWeb: decanting the web for the finest text data at scale}}.

\bibitem[{Penedo et~al.(2023)Penedo, Malartic, Hesslow, Cojocaru, Cappelli, Alobeidli, Pannier, Almazrouei, and Launay}]{penedo2023refinedweb}
Guilherme Penedo, Quentin Malartic, Daniel Hesslow, Ruxandra Cojocaru, Alessandro Cappelli, Hamza Alobeidli, Baptiste Pannier, Ebtesam Almazrouei, and Julien Launay. 2023.
\newblock \href {https://arxiv.org/abs/2306.01116} {The refinedweb dataset for falcon llm: Outperforming curated corpora with web data, and web data only}.
\newblock \emph{Preprint}, arXiv:2306.01116.

\bibitem[{Ponti et~al.(2020)Ponti, Glavas, Majewska, Liu, Vulic, and Korhonen}]{xcopa}
Edoardo~Maria Ponti, Goran Glavas, Olga Majewska, Qianchu Liu, Ivan Vulic, and Anna Korhonen. 2020.
\newblock \href {https://arxiv.org/abs/2005.00333} {{XCOPA:} {A} multilingual dataset for causal commonsense reasoning}.
\newblock \emph{CoRR}, abs/2005.00333.

\bibitem[{Radford et~al.(2019)Radford, Wu, Child, Luan, Amodei, and Sutskever}]{gpt2-radford2019language}
Alec Radford, Jeff Wu, Rewon Child, David Luan, Dario Amodei, and Ilya Sutskever. 2019.
\newblock Language models are unsupervised multitask learners.

\bibitem[{Rae et~al.(2022)Rae, Borgeaud, Cai, Millican, Hoffmann, Song, Aslanides, Henderson, Ring, Young, Rutherford, Hennigan, Menick, Cassirer, Powell, van~den Driessche, Hendricks, Rauh, Huang, Glaese, Welbl, Dathathri, Huang, Uesato, Mellor, Higgins, Creswell, McAleese, Wu, Elsen, Jayakumar, Buchatskaya, Budden, Sutherland, Simonyan, Paganini, Sifre, Martens, Li, Kuncoro, Nematzadeh, Gribovskaya, Donato, Lazaridou, Mensch, Lespiau, Tsimpoukelli, Grigorev, Fritz, Sottiaux, Pajarskas, Pohlen, Gong, Toyama, de~Masson~d'Autume, Li, Terzi, Mikulik, Babuschkin, Clark, de~Las~Casas, Guy, Jones, Bradbury, Johnson, Hechtman, Weidinger, Gabriel, Isaac, Lockhart, Osindero, Rimell, Dyer, Vinyals, Ayoub, Stanway, Bennett, Hassabis, Kavukcuoglu, and Irving}]{rae2022scaling}
Jack~W. Rae, Sebastian Borgeaud, Trevor Cai, Katie Millican, Jordan Hoffmann, Francis Song, John Aslanides, Sarah Henderson, Roman Ring, Susannah Young, Eliza Rutherford, Tom Hennigan, Jacob Menick, Albin Cassirer, Richard Powell, George van~den Driessche, Lisa~Anne Hendricks, Maribeth Rauh, Po-Sen Huang, Amelia Glaese, Johannes Welbl, Sumanth Dathathri, Saffron Huang, Jonathan Uesato, John Mellor, Irina Higgins, Antonia Creswell, Nat McAleese, Amy Wu, Erich Elsen, Siddhant Jayakumar, Elena Buchatskaya, David Budden, Esme Sutherland, Karen Simonyan, Michela Paganini, Laurent Sifre, Lena Martens, Xiang~Lorraine Li, Adhiguna Kuncoro, Aida Nematzadeh, Elena Gribovskaya, Domenic Donato, Angeliki Lazaridou, Arthur Mensch, Jean-Baptiste Lespiau, Maria Tsimpoukelli, Nikolai Grigorev, Doug Fritz, Thibault Sottiaux, Mantas Pajarskas, Toby Pohlen, Zhitao Gong, Daniel Toyama, Cyprien de~Masson~d'Autume, Yujia Li, Tayfun Terzi, Vladimir Mikulik, Igor Babuschkin, Aidan Clark, Diego de~Las~Casas, Aurelia Guy, Chris Jones,
  James Bradbury, Matthew Johnson, Blake Hechtman, Laura Weidinger, Iason Gabriel, William Isaac, Ed~Lockhart, Simon Osindero, Laura Rimell, Chris Dyer, Oriol Vinyals, Kareem Ayoub, Jeff Stanway, Lorrayne Bennett, Demis Hassabis, Koray Kavukcuoglu, and Geoffrey Irving. 2022.
\newblock \href {https://arxiv.org/abs/2112.11446} {{Scaling Language Models: Methods, Analysis \& Insights from Training Gopher}}.
\newblock \emph{Preprint}, arXiv:2112.11446.

\bibitem[{Rae et~al.(2021)Rae, Borgeaud, Cai, Millican, Hoffmann, Song, Aslanides, Henderson, Ring, Young, Rutherford, Hennigan, Menick, Cassirer, Powell, van~den Driessche, Hendricks, Rauh, Huang, Glaese, Welbl, Dathathri, Huang, Uesato, Mellor, Higgins, Creswell, McAleese, Wu, Elsen, Jayakumar, Buchatskaya, Budden, Sutherland, Simonyan, Paganini, Sifre, Martens, Li, Kuncoro, Nematzadeh, Gribovskaya, Donato, Lazaridou, Mensch, Lespiau, Tsimpoukelli, Grigorev, Fritz, Sottiaux, Pajarskas, Pohlen, Gong, Toyama, de~Masson~d’Autume, Li, Terzi, Mikulik, Babuschkin, Clark, de~Las~Casas, Guy, Jones, Bradbury, Johnson, Hechtman, Weidinger, Gabriel, Isaac, Lockhart, Osindero, Rimell, Dyer, Vinyals, Ayoub, Stanway, Bennett, Hassabis, Kavukcuoglu, and Irving}]{Rae2021Gopher}
Jack~W. Rae, Sebastian Borgeaud, Trevor Cai, Katie Millican, Jordan Hoffmann, Francis Song, John Aslanides, Sarah Henderson, Roman Ring, Susannah Young, Eliza Rutherford, Tom Hennigan, Jacob Menick, Albin Cassirer, Richard Powell, George van~den Driessche, Lisa~Anne Hendricks, Maribeth Rauh, Po-Sen Huang, Amelia Glaese, Johannes Welbl, Sumanth Dathathri, Saffron Huang, Jonathan Uesato, John Mellor, Irina Higgins, Antonia Creswell, Nat McAleese, Amy Wu, Erich Elsen, Siddhant Jayakumar, Elena Buchatskaya, David Budden, Esme Sutherland, Karen Simonyan, Michela Paganini, Laurent Sifre, Lena Martens, Xiang~Lorraine Li, Adhiguna Kuncoro, Aida Nematzadeh, Elena Gribovskaya, Domenic Donato, Angeliki Lazaridou, Arthur Mensch, Jean-Baptiste Lespiau, Maria Tsimpoukelli, Nikolai Grigorev, Doug Fritz, Thibault Sottiaux, Mantas Pajarskas, Toby Pohlen, Zhitao Gong, Daniel Toyama, Cyprien de~Masson~d’Autume, Yujia Li, Tayfun Terzi, Vladimir Mikulik, Igor Babuschkin, Aidan Clark, Diego de~Las~Casas, Aurelia Guy, Chris
  Jones, James Bradbury, Matthew Johnson, Blake Hechtman, Laura Weidinger, Iason Gabriel, William Isaac, Ed~Lockhart, Simon Osindero, Laura Rimell, Chris Dyer, Oriol Vinyals, Kareem Ayoub, Jeff Stanway, Lorrayne Bennett, Demis Hassabis, Koray Kavukcuoglu, and Geoffrey Irving. 2021.
\newblock Scaling language models: Methods, analysis \& insights from training gopher.

\bibitem[{Raffel et~al.(2020)Raffel, Shazeer, Roberts, Lee, Narang, Matena, Zhou, Li, and Liu}]{raffel2020exploring}
Colin Raffel, Noam Shazeer, Adam Roberts, Katherine Lee, Sharan Narang, Michael Matena, Yanqi Zhou, Wei Li, and Peter~J Liu. 2020.
\newblock Exploring the limits of transfer learning with a unified text-to-text transformer.
\newblock \emph{The Journal of Machine Learning Research}, 21(1):5485--5551.

\bibitem[{Raffel et~al.(2019)Raffel, Shazeer et~al.}]{t5}
Colin Raffel, Noam Shazeer, et~al. 2019.
\newblock Exploring the {Limits} of {Transfer} {Learning} with a {Unified} {Text}-to-{Text} {Transformer}.
\newblock \emph{ArXiv}, abs/1910.10683.

\bibitem[{Rogers(2021)}]{rogers2021changing}
Anna Rogers. 2021.
\newblock \href {https://arxiv.org/abs/2105.13947} {Changing the world by changing the data}.
\newblock \emph{Preprint}, arXiv:2105.13947.

\bibitem[{Sagawa et~al.(2020)Sagawa, Koh, Hashimoto, and Liang}]{sagawa2020distributionally}
Shiori Sagawa, Pang~Wei Koh, Tatsunori~B. Hashimoto, and Percy Liang. 2020.
\newblock \href {https://arxiv.org/abs/1911.08731} {Distributionally robust neural networks for group shifts: On the importance of regularization for worst-case generalization}.
\newblock \emph{Preprint}, arXiv:1911.08731.

\bibitem[{Sakaguchi et~al.(2020)Sakaguchi, Bras, Bhagavatula, and Choi}]{Sakaguchi2020WINOGRANDEAA}
Keisuke Sakaguchi, Ronan~Le Bras, Chandra Bhagavatula, and Yejin Choi. 2020.
\newblock Winogrande: An adversarial winograd schema challenge at scale.
\newblock In \emph{AAAI}.

\bibitem[{Schwenk et~al.(2019)Schwenk, Wenzek, Edunov, Grave, and Joulin}]{schwenk2019ccmatrix}
Holger Schwenk, Guillaume Wenzek, Sergey Edunov, Edouard Grave, and Armand Joulin. 2019.
\newblock Ccmatrix: Mining billions of high-quality parallel sentences on the web.
\newblock \emph{arXiv preprint arXiv:1911.04944}.

\bibitem[{Shazeer(2020)}]{shazeer2020glu}
Noam Shazeer. 2020.
\newblock \href {https://arxiv.org/abs/2002.05202} {Glu variants improve transformer}.
\newblock \emph{Preprint}, arXiv:2002.05202.

\bibitem[{Shliazhko et~al.(2022)Shliazhko, Fenogenova, Tikhonova, Mikhailov, Kozlova, and Shavrina}]{shliazhko2022mgpt}
Oleh Shliazhko, Alena Fenogenova, Maria Tikhonova, Vladislav Mikhailov, Anastasia Kozlova, and Tatiana Shavrina. 2022.
\newblock \href {https://arxiv.org/abs/2204.07580} {mgpt: Few-shot learners go multilingual}.
\newblock \emph{Preprint}, arXiv:2204.07580.

\bibitem[{Smith et~al.(2022{\natexlab{a}})Smith, Patwary, Norick, LeGresley, Rajbhandari, Casper, Liu, Prabhumoye, Zerveas, Korthikanti, Zhang, Child, Aminabadi, Bernauer, Song, Shoeybi, He, Houston, Tiwary, and Catanzaro}]{smith2022using}
Shaden Smith, Mostofa Patwary, Brandon Norick, Patrick LeGresley, Samyam Rajbhandari, Jared Casper, Zhun Liu, Shrimai Prabhumoye, George Zerveas, Vijay Korthikanti, Elton Zhang, Rewon Child, Reza~Yazdani Aminabadi, Julie Bernauer, Xia Song, Mohammad Shoeybi, Yuxiong He, Michael Houston, Saurabh Tiwary, and Bryan Catanzaro. 2022{\natexlab{a}}.
\newblock \href {https://arxiv.org/abs/2201.11990} {Using deepspeed and megatron to train megatron-turing nlg 530b, a large-scale generative language model}.
\newblock \emph{Preprint}, arXiv:2201.11990.

\bibitem[{Smith et~al.(2022{\natexlab{b}})Smith, Patwary, Norick, LeGresley, Rajbhandari, Casper, Liu, Prabhumoye, Zerveas, Korthikanti, Zheng, Child, Aminabadi, Bernauer, Song, Shoeybi, He, Houston, Tiwary, and Catanzaro}]{smith2022}
Shaden Smith, Mostofa Patwary, Brandon Norick, Patrick LeGresley, Samyam Rajbhandari, Jared Casper, Zhun Liu, Shrimai Prabhumoye, George Zerveas, Vijay Korthikanti, Elton Zheng, Rewon Child, Reza~Yazdani Aminabadi, Julie Bernauer, Xia Song, Mohammad Shoeybi, Yuxiong He, Michael Houston, Saurabh Tiwary, and Bryan Catanzaro. 2022{\natexlab{b}}.
\newblock \href {https://arxiv.org/abs/2201.11990} {{Using DeepSpeed and Megatron to Train Megatron-Turing {NLG} 530B, {A} Large-Scale Generative Language Model}}.
\newblock \emph{CoRR}, abs/2201.11990.

\bibitem[{Soldaini et~al.(2024)Soldaini, Kinney, Bhagia, Schwenk, Atkinson, Authur, Bogin, Chandu, Dumas, Elazar, Hofmann, Jha, Kumar, Lucy, Lyu, Lambert, Magnusson, Morrison, Muennighoff, Naik, Nam, Peters, Ravichander, Richardson, Shen, Strubell, Subramani, Tafjord, Walsh, Zettlemoyer, Smith, Hajishirzi, Beltagy, Groeneveld, Dodge, and Lo}]{soldaini2024dolma}
Luca Soldaini, Rodney Kinney, Akshita Bhagia, Dustin Schwenk, David Atkinson, Russell Authur, Ben Bogin, Khyathi Chandu, Jennifer Dumas, Yanai Elazar, Valentin Hofmann, Ananya~Harsh Jha, Sachin Kumar, Li~Lucy, Xinxi Lyu, Nathan Lambert, Ian Magnusson, Jacob Morrison, Niklas Muennighoff, Aakanksha Naik, Crystal Nam, Matthew~E. Peters, Abhilasha Ravichander, Kyle Richardson, Zejiang Shen, Emma Strubell, Nishant Subramani, Oyvind Tafjord, Pete Walsh, Luke Zettlemoyer, Noah~A. Smith, Hannaneh Hajishirzi, Iz~Beltagy, Dirk Groeneveld, Jesse Dodge, and Kyle Lo. 2024.
\newblock \href {https://arxiv.org/abs/2402.00159} {Dolma: an open corpus of three trillion tokens for language model pretraining research}.
\newblock \emph{Preprint}, arXiv:2402.00159.

\bibitem[{Su et~al.(2023)Su, Lu, Pan, Murtadha, Wen, and Liu}]{su2023roformer}
Jianlin Su, Yu~Lu, Shengfeng Pan, Ahmed Murtadha, Bo~Wen, and Yunfeng Liu. 2023.
\newblock \href {https://arxiv.org/abs/2104.09864} {Roformer: Enhanced transformer with rotary position embedding}.
\newblock \emph{Preprint}, arXiv:2104.09864.

\bibitem[{Team(2024{\natexlab{a}})}]{geminiteam2024geminipro1.5}
Gemini Team. 2024{\natexlab{a}}.
\newblock \href {https://arxiv.org/abs/2403.05530} {Gemini 1.5: Unlocking multimodal understanding across millions of tokens of context}.
\newblock \emph{Preprint}, arXiv:2403.05530.

\bibitem[{Team(2024{\natexlab{b}})}]{geminiteam2024gemini}
Gemini Team. 2024{\natexlab{b}}.
\newblock \href {https://arxiv.org/abs/2312.11805} {Gemini: A family of highly capable multimodal models}.
\newblock \emph{Preprint}, arXiv:2312.11805.

\bibitem[{Team et~al.(2024)Team, Ormazabal, Zheng, de~Masson~d'Autume, Yogatama, Fu, Ong, Chen, Lamprecht, Pham, Ong, Aleksiev, Li, Henderson, Bain, Artetxe, Relan, Padlewski, Liu, Chen, Phua, Yang, Tay, Wang, Zhu, and Xie}]{rekateam2024reka}
Reka Team, Aitor Ormazabal, Che Zheng, Cyprien de~Masson~d'Autume, Dani Yogatama, Deyu Fu, Donovan Ong, Eric Chen, Eugenie Lamprecht, Hai Pham, Isaac Ong, Kaloyan Aleksiev, Lei Li, Matthew Henderson, Max Bain, Mikel Artetxe, Nishant Relan, Piotr Padlewski, Qi~Liu, Ren Chen, Samuel Phua, Yazheng Yang, Yi~Tay, Yuqi Wang, Zhongkai Zhu, and Zhihui Xie. 2024.
\newblock \href {https://arxiv.org/abs/2404.12387} {Reka core, flash, and edge: A series of powerful multimodal language models}.
\newblock \emph{Preprint}, arXiv:2404.12387.

\bibitem[{Thirumuruganathan et~al.(2020)Thirumuruganathan, Tang, Ouzzani, and Doan}]{Thirumuruganathan2020DataCW}
Saravanan Thirumuruganathan, Nan Tang, Mourad Ouzzani, and AnHai Doan. 2020.
\newblock \href {https://api.semanticscholar.org/CorpusID:214613368} {Data curation with deep learning}.
\newblock In \emph{International Conference on Extending Database Technology}.

\bibitem[{Touvron et~al.(2023{\natexlab{a}})Touvron, Lavril, Izacard, Martinet, Lachaux, Lacroix, Rozière, Goyal, Hambro, Azhar, Rodriguez, Joulin, Grave, and Lample}]{touvron2023llama}
Hugo Touvron, Thibaut Lavril, Gautier Izacard, Xavier Martinet, Marie-Anne Lachaux, Timothée Lacroix, Baptiste Rozière, Naman Goyal, Eric Hambro, Faisal Azhar, Aurelien Rodriguez, Armand Joulin, Edouard Grave, and Guillaume Lample. 2023{\natexlab{a}}.
\newblock \href {https://arxiv.org/abs/2302.13971} {Llama: Open and efficient foundation language models}.
\newblock \emph{Preprint}, arXiv:2302.13971.

\bibitem[{Touvron et~al.(2023{\natexlab{b}})Touvron, Martin, Stone, Albert, Almahairi, Babaei, Bashlykov, Batra, Bhargava, Bhosale et~al.}]{touvron2023llama2}
Hugo Touvron, Louis Martin, Kevin Stone, Peter Albert, Amjad Almahairi, Yasmine Babaei, Nikolay Bashlykov, Soumya Batra, Prajjwal Bhargava, Shruti Bhosale, et~al. 2023{\natexlab{b}}.
\newblock {Llama 2: Open Foundation and Fine-tuned Chat Models}.
\newblock \emph{arXiv preprint arXiv:2307.09288}.

\bibitem[{Trinh and Le(2018)}]{DBLP:journals/corr/abs-1806-02847}
Trieu~H. Trinh and Quoc~V. Le. 2018.
\newblock \href {https://arxiv.org/abs/1806.02847} {A simple method for commonsense reasoning}.
\newblock \emph{CoRR}, abs/1806.02847.

\bibitem[{Vaswani et~al.(2017)Vaswani, Shazeer, Parmar, Uszkoreit, Jones, Gomez, Kaiser, and Polosukhin}]{Vaswani+2017}
Ashish Vaswani, Noam Shazeer, Niki Parmar, Jakob Uszkoreit, Llion Jones, Aidan~N Gomez, \L~ukasz Kaiser, and Illia Polosukhin. 2017.
\newblock \href {https://proceedings.neurips.cc/paper_files/paper/2017/file/3f5ee243547dee91fbd053c1c4a845aa-Paper.pdf} {Attention is all you need}.
\newblock In \emph{Advances in Neural Information Processing Systems}, volume~30. Curran Associates, Inc.

\bibitem[{Wang et~al.(2019)Wang, Shi, Kim, Oh, Yang, Zhang, and Yu}]{wang2019persuasion}
Xuewei Wang, Weiyan Shi, Richard Kim, Yoojung Oh, Sijia Yang, Jingwen Zhang, and Zhou Yu. 2019.
\newblock Persuasion for good: Towards a personalized persuasive dialogue system for social good.
\newblock \emph{arXiv preprint arXiv:1906.06725}.

\bibitem[{Welbl et~al.(2021)Welbl, Glaese, Uesato, Dathathri, Mellor, Hendricks, Anderson, Kohli, Coppin, and Huang}]{welbl2021challenges}
Johannes Welbl, Amelia Glaese, Jonathan Uesato, Sumanth Dathathri, John Mellor, Lisa~Anne Hendricks, Kirsty Anderson, Pushmeet Kohli, Ben Coppin, and Po-Sen Huang. 2021.
\newblock Challenges in detoxifying language models.
\newblock \emph{arXiv preprint arXiv:2109.07445}.

\bibitem[{Wu et~al.(2023)Wu, Irsoy, Lu, Dabravolski, Dredze, Gehrmann, Kambadur, Rosenberg, and Mann}]{wu2023bloomberggpt}
Shijie Wu, Ozan Irsoy, Steven Lu, Vadim Dabravolski, Mark Dredze, Sebastian Gehrmann, Prabhanjan Kambadur, David Rosenberg, and Gideon Mann. 2023.
\newblock \href {https://arxiv.org/abs/2303.17564} {Bloomberggpt: A large language model for finance}.
\newblock \emph{Preprint}, arXiv:2303.17564.

\bibitem[{Xie et~al.(2023{\natexlab{a}})Xie, Pham, Dong, Du, Liu, Lu, Liang, Le, Ma, and Yu}]{xie2023doremi}
Sang~Michael Xie, Hieu Pham, Xuanyi Dong, Nan Du, Hanxiao Liu, Yifeng Lu, Percy Liang, Quoc~V. Le, Tengyu Ma, and Adams~Wei Yu. 2023{\natexlab{a}}.
\newblock \href {https://arxiv.org/abs/2305.10429} {Doremi: Optimizing data mixtures speeds up language model pretraining}.
\newblock \emph{Preprint}, arXiv:2305.10429.

\bibitem[{Xie et~al.(2023{\natexlab{b}})Xie, Santurkar, Ma, and Liang}]{xie2023data}
Sang~Michael Xie, Shibani Santurkar, Tengyu Ma, and Percy Liang. 2023{\natexlab{b}}.
\newblock \href {https://arxiv.org/abs/2302.03169} {Data selection for language models via importance resampling}.
\newblock \emph{Preprint}, arXiv:2302.03169.

\bibitem[{Xu et~al.(2021)Xu, Pathak, Wallace, Gururangan, Sap, and Klein}]{xu2021detoxifying}
Albert Xu, Eshaan Pathak, Eric Wallace, Suchin Gururangan, Maarten Sap, and Dan Klein. 2021.
\newblock \href {https://arxiv.org/abs/2104.06390} {Detoxifying language models risks marginalizing minority voices}.
\newblock \emph{Preprint}, arXiv:2104.06390.

\bibitem[{Xue et~al.(2020)Xue, Constant, Roberts, Kale, Al-Rfou, Siddhant, Barua, and Raffel}]{mt5}
Linting Xue, Noah Constant, Adam Roberts, Mihir Kale, Rami Al-Rfou, Aditya Siddhant, Aditya Barua, and Colin Raffel. 2020.
\newblock mt5: A massively multilingual pre-trained text-to-text transformer.
\newblock \emph{arXiv preprint arXiv:2010.11934}.

\bibitem[{Yang et~al.(2020)Yang, Dai, Yang, Carbonell, Salakhutdinov, and Le}]{yang2020xlnet}
Zhilin Yang, Zihang Dai, Yiming Yang, Jaime Carbonell, Ruslan Salakhutdinov, and Quoc~V. Le. 2020.
\newblock \href {https://arxiv.org/abs/1906.08237} {Xlnet: Generalized autoregressive pretraining for language understanding}.
\newblock \emph{Preprint}, arXiv:1906.08237.

\bibitem[{Ye et~al.(2023)Ye, Tao, and Kong}]{ye2023language}
Jiacheng Ye, Xijia Tao, and Lingpeng Kong. 2023.
\newblock \href {https://arxiv.org/abs/2306.06688} {Language versatilists vs. specialists: An empirical revisiting on multilingual transfer ability}.
\newblock \emph{Preprint}, arXiv:2306.06688.

\bibitem[{Zellers et~al.(2019)Zellers, Holtzman, Bisk, Farhadi, and Choi}]{Zellers2019HellaSwagCA}
Rowan Zellers, Ari Holtzman, Yonatan Bisk, Ali Farhadi, and Yejin Choi. 2019.
\newblock Hellaswag: Can a machine really finish your sentence?
\newblock In \emph{ACL}.

\bibitem[{Zhang et~al.(2022)Zhang, Roller, Goyal, Artetxe, Chen, Chen, Dewan, Diab, Li, Lin et~al.}]{zhang2022opt}
Susan Zhang, Stephen Roller, Naman Goyal, Mikel Artetxe, Moya Chen, Shuohui Chen, Christopher Dewan, Mona Diab, Xian Li, Xi~Victoria Lin, et~al. 2022.
\newblock Opt: Open pre-trained transformer language models.
\newblock \emph{arXiv preprint arXiv:2205.01068}.

\bibitem[{Zhou and Choi(2018)}]{zhou2018they}
Ethan Zhou and Jinho~D Choi. 2018.
\newblock They exist! introducing plural mentions to coreference resolution and entity linking.
\newblock In \emph{Proceedings of the 27th International Conference on Computational Linguistics}, pages 24--34.

\end{thebibliography}

%\clearpage
%\newpage

\appendix

\section{Data Sources}
\label{sec:appendix_data_sources}

\subsection{English Data Sources}
\label{sec:appendix_data_sources_english}

Table \ref{tab:english_data_summary} shares the datasets which compose our English corpus. We share further detail on how we gathered the datasets from each category.

\begin{table}[h]
\centering
\adjustbox{max width=\textwidth}{%
 \begin{tabular}{l l c }
    \toprule
    \textbf{Data source} & \textbf{Dataset name} & 
    \textbf{Tokens (B)} \\
    \midrule  
    \multirow{7}{*}{Web Crawl} & 
     CC 2022-40 & 284.3  \\
     & Re-crawled C4 & 174.8   \\
     & CC 2019-35 & 165.1  \\
     & CC 2020-50 & 141.9  \\
     & CC 2021-04  & 68.2  \\
     & Pile-CC &  41.2 \\
     & OpenWebText2   & 14.0   \\
    \midrule
    \multirow{1}{*}{News} & 
    CC NEWS & 94.2  \\
    \midrule
    \multirow{2}{*}{Misc} & 
    ROOTS & 104.5  \\
    & Wikipedia & 4.3  \\
    \midrule
    \multirow{1}{*}{Conv.} & 
    Reddit + others  & 59.1   \\
    \midrule
    \multirow{4}{*}{Books} & 
    Books3 &  25.1   \\
    & Stories & 5.3  \\
    & Gutenberg  & 2.5  \\
    & BookCorpus2  & 1.5 \\
    \midrule
    \multirow{4}{*}{Scientific} & 
    ArXiv  & 18.7   \\
    & StackExchange & 9.8 \\
    & PubMed Abstracts & 4.2   \\
    & NIH ExPorter & 0.3  \\
    \bottomrule
  \end{tabular}}
  \caption{\label{tab:english_data_summary} 
  Summary of each of the datasets that make up our English corpus}
  \label{tab:english_data_summary}
\end{table}

\paragraph{Web Crawl}
To acquire a significant amount of web-crawl data, we downloaded all Common Crawl web archive
(WARC) files originating from the CC-2020-50, CC-2019-35, CC-2021-04 and CC-2022-40 snapshots.
Additionally, we re-crawled all URLs provided by the documents within the C4 corpus \citep{t5}. 
While many of these URLs were no longer active, we were able to re-crawl approximately 
1.7 TB of web pages contained within the C4 dataset. 
To add to our collected web-crawl corpus, we also used the 
pre-preprocessed documents available within Pile-CC \citep{pile-dataset-2020}.

\paragraph{News}
To curate our news dataset, we downloaded all Common Crawl News WARC files between 2016 and 
October 2022. 

\paragraph{Conversational}
Our conversational dataset was constructed primarily from the Pushshift Reddit dataset \citep{reddit-dataset-2020}, with small amounts of other public datasets such as  CaSiNo \citep{chawla2021casino}, 
Wikipedia Talk Pages \citep{ferschke2012behind}, Persuasion for good \citep{wang2019persuasion},  
Friends \citep{zhou2018they}, Chromium, \citep{meyers2018dataset} and Ubuntu dialogue conversational datasets \citep{lowe2015ubuntu}. 

The Reddit dataset was pre-processed to ensure that only the longest conversation thread is sampled per post to avoid duplicate text that can arise from sampling many or all possible conversation subtrees \citep{zhang2022opt}. Reddit usernames are anonymized with random alphanumeric strings while preserving speaker information within the conversation. Given the prevalence of toxic and harmful content on Reddit, we filter out conversations that have a toxicity score $>= 0.5$ according to Perspective API\footnote{\url{https://perspectiveapi.com/}}.

\paragraph{Books}
Our books dataset consisted of documents originating from the Books3, Gutenberg (PG-19), BookCorpus2 (all provided by the Pile), as well as documents from the CC-Stories dataset \citep{DBLP:journals/corr/abs-1806-02847}.

\paragraph{Scientific}
We curated all scientific documents from sub-datasets contained within the Pile. Specifically, we used the StackExchange, PubMed Abstracts, NIH Exporter and ArXiv datasets.

\paragraph{Misc}
As a miscellaneous category, we lump together the Wikipedia and ROOTS \citep{bigscience-dataset-2022} datasets.

\subsection{Multilingual Data Sources}
\label{sec:appendix_data_sources_multilingual}

Our multilingual dataset consists of 52 languages, 50 of which were curated from the 
CC-2022-40 Common Crawl snapshot. For Chinese and Japanese, we used documents from the mC4 corpus \citep{mt5}. This was a consequence of the inability of our text extraction library to parse languages without spacing. 
Table \ref{tab:ml_data_summary} provides a summary of the
multilingual web crawl data that made up our multingual corpus.

Additionally, we used an English-centric sentence-level parallel corpus of 32 languages (Details in Table.\ref{tab:nmt_composition}). This was  collected largely from data sources such as CC-Matrix \citep{schwenk2019ccmatrix}, CC-Aligned \citep{el2019ccaligned} and Paracrawl \citep{espla2019paracrawl}. Multiple examples are formatted into a document using few-shot templates with the number of in-context examples from 0-10 following an exponentially decaying probability of selection. 

\begin{table*}
\centering
  \begin{tabular}{cr|cr|cr|cr}
    \toprule
    \textbf{ISO} & \textbf{Tokens (B)} &
    \textbf{ISO} & \textbf{Tokens (B)} &
    \textbf{ISO} & \textbf{Tokens (B)} &
    \textbf{ISO} & \textbf{Tokens (B)} \\
    \midrule
     RU  & 94.52  & FA  & 6.59  & HI  & 2.60 & IS  & 0.38  \\
     JA  & 70.52 & RO  & 6.58  & SK  & 2.58 & UR  & 0.37  \\
     DE  & 48.98  & TR  & 6.46  & HR  & 2.45 & AZ  & 0.37  \\
     ES  & 46.50   & EL  & 6.43   & CA  & 2.12 & MR  & 0.33 \\
     FR  & 44.30   & SV  & 6.39 & LT   & 1.69 & KA  & 0.32 \\
     ZH  & 43.41   & HU  & 5.89   & HE  & 1.47 & MK  & 0.32 \\
     IT  & 26.40   & AR  & 5.74   & SL  & 1.33  & NE  & 0.31 \\
     NL  & 15.64   & NO  & 5.61 & SR  & 1.24 & KK  & 0.30   \\
     VI  & 15.16   & FI  & 4.11 & ET  & 1.24 & HY  & 0.29   \\
     PL  & 14.50   & DA  & 3.79 & BN  & 0.90 & GL  & 0.29  \\
     PT  & 11.99   & UK  & 3.63 & LV  & 0.84 & ML  & 0.25  \\
     ID  & 10.90   & BG  & 3.37 & TA  & 0.82 & TE  & 0.24 \\
     CS  & 7.23   & KO  & 3.05 &   SQ  & 0.49 & KN  & 0.18  \\
    \bottomrule
  \end{tabular}
   \caption{\label{tab:ml_data_summary} Summary of our multilingual web crawl data consisting of 52 languages. All languages except for
    JA and ZH were curated from the 2022-40 CC snapshot. The JA and ZH
    data were curated from the mC4 corpus.}
\end{table*}

\begin{table*}
\centering
\adjustbox{max width=\textwidth}{%
\begin{tabular}{cccccccc}\toprule
\multirow{2}{*}{\textbf{Language}} & \textbf{Percentage} & \multirow{2}{*}{\textbf{Language}} & \textbf{Percentage} & \multirow{2}{*}{\textbf{Language}} & \textbf{Percentage}  & \multirow{2}{*}{\textbf{Language}} & \textbf{Percentage} \\
& \textbf{(\%)}  & & \textbf{(\%)} & & \textbf{(\%)} & & \textbf{(\%)} \\ \midrule
Spanish & 12.84 & Indonesian & 3.12 & Japanese & 2.30 & Lithuanian & 1.39 \\
French & 10.52 & Portuguese & 2.90 & Norwegian & 2.19 & Bulgarian & 1.30 \\
German & 9.78 & Polish & 2.88 & Hungarian & 2.13 & Hindi & 1.17 \\
Italian & 5.48 & Czech & 2.74 & Ukrainian & 1.90 & Slovak & 0.99 \\
Russian & 5.25 & Turkish & 2.60 & Finnish & 1.84 & Slovenian & 0.91 \\
Dutch & 4.81 & Vietnamese & 2.54 & Swedish & 1.73 & Estonian & 0.81 \\
Chinese & 3.61 & Greek & 2.39 & Korean & 1.54 & Latvian & 0.76 \\
Arabic & 3.20 & Romanian & 2.32 & Danish & 1.53 & Croatian & 0.55 \\ \bottomrule
%Spanish & 12.84 & 128390582 & Indonesian & 3.12 & 31222550 & Japanese & 2.30 & 23042564 & Lithuanian & 1.39 & 13945583 \\
%French & 10.52 & 105244340 & Portuguese & 2.90 & 28981313 & Norwegian & 2.19 & 21932524 & Bulgarian & 1.30 & 13045558 \\
%German & 9.78 & 97766621 & Polish & 2.88 & 28764416 & Hungarian & 2.13 & 21275350 & Hindi & 1.17 & 11712895 \\
%Italian & 5.48 & 54756146 & Czech & 2.74 & 27358621 & Ukrainian & 1.90 & 18958463 & Slovak & 0.99 & 9870182 \\
%Russian & 5.25 & 52549618 & Turkish & 2.60 & 25995971 & Finnish & 1.84 & 18365964 & Slovenian & 0.91 & 9068199 \\
%Dutch & 4.81 & 48141829 & Vietnamese & 2.54 & 25389772 & Swedish & 1.73 & 17275712 & Estonian & 0.81 & 8104915 \\
%Chinese & 3.61 & 36136238 & Greek & 2.39 & 23902451 & Korean & 1.54 & 15374503 & Latvian & 0.76 & 7555638 \\
%Arabic & 3.20 & 31977273 & Romanian & 2.32 & 23159372 & Danish & 1.53 & 15253052 & Croatian & 0.55 & 5481770 \\
\end{tabular}}
\caption{\label{tab:nmt_composition}The language composition of our parallel machine translation corpus.}
\end{table*}

\subsection{Code Data Sources}
\label{sec:appendix_data_sources_code}

Our source code dataset was mainly constructed from a subset of the Stack v1.2 dataset \citep{kocetkov2022stack}. Table \ref{tab:code_data_summary} list the selected languages and their token counts. While the dataset is distributed with each file as a single document, we pre-process the data further to concatenate all files of a particular language from a repository into a single long document to allow the model to attend across files.

\begin{table*}
  \centering
  \begin{tabular}{lr|lr|lr}
    \toprule
    \textbf{Language} & \textbf{Tokens (B)} &
    \textbf{Language} & \textbf{Tokens (B)} &
    \textbf{Language} & \textbf{Tokens (B)} \\
    \midrule
    Javascript & 21.12 & Rust & 2.81 & Pascal & 0.68 \\
    Markdown & 20.27 & Jupyter & 2.58 & Assembly & 0.67 \\
    Java & 19.84 & Ruby & 2.29 & Fortran & 0.65 \\
    Python & 19.49 & Swift & 2.02 & Makefile  & 0.54  \\
    PHP & 18.87 & JSON & 1.78 & Julia & 0.52  \\
    C & 18.26 & \TeX & 1.76  & Mathematica & 0.51  \\
    C++ & 15.79 & Scala & 1.29 & Visual Basic & 0.42 \\
    C\#  & 12.05 & YAML & 1.28 & VHDL & 0.42  \\
    Go & 9.03 & Shell & 1.18 & Common Lisp & 0.24 \\
    HTML & 8.97 & Dart & 1.08 & Cuda & 0.21  \\
    Typescript & 8.16 & Lua & 1.00 & System Verilog & 0.16 \\

    SQL & 5.31 & reStructuredText & 0.96 & Docker & 0.16 \\
    CSS & 4.96 & Perl & 0.83 & Omniverse & 0.03  \\
    XML & 2.97 & Haskell & 0.72 &  &  \\
    
    % Scala & 0.71  & 1.88  & 0.38  & 0.065 \\
    % YAML  & 0.59  & 1.62  & 0.36  & 0.053 \\
    % Shell & 0.48  & 1.37  & 0.35  & 0.043 \\
    % Dart  & 0.46  & 1.33  & 0.34  & 0.042 \\
    % Lua & 0.35  & 1.09  & 0.32  & 0.032 \\
    % Perl  & 0.27  & 0.90  & 0.30  & 0.025 \\
    % CUDA  & 0.25  & 0.26  & 0.96  & 0.022 \\
    % Haskell & 0.25  & 0.83  & 0.30  & 0.022 \\
    % \TeX & 0.22  & 0.77  & 0.29  & 0.020 \\
    % Makefile  & 0.21  & 0.71  & 0.29  & 0.019 \\
    % Fortran & 0.18  & 0.64  & 0.27  & 0.016 \\
    % Common Lisp & 0.16 &  0.59 & 0.27 & 0.014 \\
    % Julia & 0.14  & 0.53  & 0.26  & 0.012 \\
    % RST & 0.13  & 0.52  & 0.26  & 0.012 \\
    % Visual Basic &  0.12 &  0.49 &  0.25 &  0.011 \\
    % Pascal  & 0.11  & 0.46  & 0.25  & 0.010 \\
    % Assembly  & 0.05  & 0.25  & 0.21  & 0.005 \\
    % Dockerfile  & 0.01  & 0.05  & 0.13  & 0.001 \\
    \bottomrule
  \end{tabular}
  \caption{\label{tab:code_data_summary} Summary of our source code corpus consisting of 41 different programming languages all of which, except for omniverse, were curated from the Stack v1.2 dataset.}
\end{table*}

\begin{table*}[!h]
\centering
\adjustbox{max width=\textwidth}{%
\begin{tabular}{l c c}\toprule
\multicolumn{1}{c}{Heuristic} & \multicolumn{1}{c}{Threshold} & \multicolumn{1}{c}{English Only}  \\ \midrule
N-gram LM Perplexity & 5000 & Yes \\ \midrule
Fraction of non-alpha-numeric characters & 0.25 & Yes \\
Fraction of words without alphabets & 0.20 & Yes \\
Fraction of numbers (in characters) & 0.15 & \\
Fraction of URLs (in characters) & 0.20 & \\
Fraction of lines starting with bullets & 0.90 & \\
Fraction of whitespaces (in characters) & 0.25 & \\
Fraction of parentheses (in characters) & 0.10 & \\
The ratio of symbols to words & 0.10 & \\ \midrule
Contains a word \textgreater 1000 characters & 1.0  & \\
Contains \textless 50 or \textgreater 100k words & 1.0  & \\ 
Contains less than 2 common English words & 1.0  & Yes\\
Mean word length \textless 3 or \textgreater 10 characters & 1.0 & \\ \midrule
Fraction of boilerplate content (in characters) & 0.40 & \\ \midrule
Duplicate line fraction & 0.30 & \\
Duplicate paragraph fraction & 0.30 & \\
Duplicate lines (by character fraction) & 0.20 & \\
Duplicate paragraph (by character fraction) & 0.10 & \\ \midrule
Repeating top n-gram fraction & 0.20 & \\
Repeating duplicate n-gram fraction & 0.20 & \\
Fraction of lines that do not end with punctuation & 0.85 & \\
Fraction of lines that end with ellipsis & 0.30 & \\ \midrule
Documents containing Pornographic content in URLs & 1.00 & \\ \bottomrule
\end{tabular}}
\caption{\label{tab:filtering_heuristics}A list of document-level data filtering heuristics and thresholds. Heuristics are borrowed or derived from \citet{Rae2021Gopher} and C4's cleaning heuristics \citep{raffel2020exploring}}
\end{table*}

\begin{table*}[!h]
\centering
\begin{tabular}{l r r}\toprule
\multicolumn{1}{c}{Heuristic} & \multicolumn{1}{c}{Min. Threshold} & \multicolumn{1}{c}{Max Threshold}\\ \midrule
Fraction of comments (in characters)&  0.001 & 0.85 \\
Number of lines of code & 5 & 20,000\\
Ratio of characters to tokens & 2 & -  \\ \bottomrule
\end{tabular}
\caption{\label{tab:code_filtering_heuristics}A list of file-level data filtering heuristics and thresholds applied to the source code data. Heuristics follow those described in \cite{allal2023santacoder}.}
\end{table*}

\section{Data Attribute Classifiers}
\label{sec:appendix_data_attribute_classifiers}
We detail the training methodology, output labels, and public release plan for each of our data attribute classifiers.

\subsection{Toxicity Classifier}

Solutions, like Perspective API, exist for quantifying the toxicity of a given piece of text. However, due to low rate limits it would be intractable to scale across the billions of documents that exist across all CC snapshots. In developing our own toxicity classifier, we aim to recapitulate the performance of Perspective API and reliably mark text which contain obscene language, threats, insults, and identity-based hate speech as having high toxicity. As a training set, we use 320K examples sourced from the Jigsaw\footnote{https://www.kaggle.com/c/jigsaw-toxic-comment-classification-challenge/overview} and Jigsaw Unintended\footnote{https://www.kaggle.com/competitions/jigsaw-unintended-bias-in-toxicity-classification} datasets. We obtain our final toxicity classifier by fine-tuning a DeBERTaV3 base model for 1 epoch on this data. The output for our toxicity classifier is a probability from 0 to 1 on whether or not a given piece of text contains toxic content. 

We evaluate our toxicity classifier by measuring its correlation with Perspective API scores on a set of 50k documents from CC. We find that the classifier obtains a Pearson correlation of 0.8 which indicates high agreement between the models. Additionally, we ask a set of human annotators to label 500 documents with toxicity scores. On this held-out test set, we find that our toxicity classifier achieves an AUC-ROC of 0.83 while Perspective API attains an AUC-ROC of 0.85. We plan to publicly release our toxicity classifier shortly. 

\subsection{Domain Classifier}
We train a domain classifier to label the content domain of a given piece of text into one of 27 potential classes: Adult, Arts and Entertainment, Autos and Vehicles, Beauty and Fitness, Books and Literature, Business and Industrial, Computers and Electronics, Finance, Food and Drink, Games, Health, Hobbies and Leisure, Home and Garden, Internet and Telecom, Jobs and Education, Law and Government, News, Online Communities, People and Society, Pets and Animals, Real Estate, Reference, Science, Sensitive Subjects, Shopping, Sports, and Travel and Transportation. The training data consists of 1 million CC documents which are labeled using Google Cloud’s Natural Language API\footnote{https://cloud.google.com/natural-language/docs/classifying-text} and 500k Wikipedia articles that are curated using the Wikipedia-API \footnote{https://pypi.org/project/Wikipedia-API/}. We train a DeBERTaV3 on two epochs of this training set. We ask a set of human annotators to label 500 held-out CC documents and evaluate both the Google API and our domain classifier on this test set. We find that our trained domain classifier matches the performance of the Google API as it achieves an accuracy of 77.9\% while the Google API achieves 77.5\%. Additionally, we publicly release our domain classifier on HuggingFace \footnote{https://huggingface.co/nvidia/domain-classifier}.

\subsection{Quality Classifier}
We train a quality classifier to label a given piece of text as either high, medium, or low quality. The training data consists of 25k CC examples that are labeled by 3 Surge AI\footnote{https://www.surgehq.ai/} annotators. We ensure that all of these annotated documents had at least greater than 2 annotators in agreement on the quality label. In these annotations we provide the following definitions of each quality class to the annotators: 
\paragraph{High}
Text which is grammatically correct, well-written, coherent between sentences and paragraphs, and without any missing punctuations or without any incomplete sentences. It also does not include any boilerplate text and has useful content.
\paragraph{Medium}
This text is mostly grammatically correct with minor errors. It may not be coherent throughout and can jump from topic to topic. It should not have many missing punctuations or incomplete sentences. It should not include a lot of boilerplate text and more than 50\% of the text should be useful.
\paragraph{Low}
This category includes text which is not grammatical, not coherent at all, or contains a lot of missing punctuations, poor capitalization of words and incomplete sentences or abrupt paragraph breaks. If the text contains pornographic content, lewd or profane language or toxic content of any kind then it is de facto low quality. Text which has a lot of boilerplate content making more than 50\% of the text useless should also be marked as “Low”.

We train a DeBERTaV3 model on this training set and find that on a held-out test set of 23k additionally labeled examples, it achieves an accuracy of 83\%. We plan to publicly release our quality classifier shortly.

\subsection{Type of Speech Classifier}
We train a type of speech classifier to label a given document into one of the following 11 document types: conversational, news, online comments, books and literature, blogs, analytical exposition (persuasive text), explanatory articles, reviews, product/company/organization/personal websites, boilerplate content, and miscellaneous. The training data consists of the same 25k CC examples labeled by 3 Surge AI annotators as the quality classifier training set. We ensure that all of these annotated documents had at least greater than 2 annotators in agreement on the type of speech label. In these annotations we provide the following definitions of each type of speech label to the annotators: 

\paragraph{Conversational}
Is this text a conversation between two or more people? Does this piece of text sound like a response to something which is not mentioned in the document? If the answer to either of the questions is “Yes” then mark the document as belonging to this category. Conversations include podcast transcripts, talk show transcripts or if there is an exchange of thoughts, feelings, ideas or information between two or more people.
\paragraph{News}
News is a form of communication that informs the public of current events, issues, and trends in society.
\paragraph{Online Comments}
Comments are messages posted by users in reaction to social media or blog posts. They can take the shape of feedback, questions, praise, or even disagreements. Comment is a short-form type of content or message that gets published on social media platforms or other online communities. You may have to check the URL of the document to get a sense of the context of the text. This category encompasses social media comments, comments in online communities, and comments on an article or a blog.
\paragraph{Books and Literature}
Is the piece of text long and seems to span multiple pages? Does it have different chapters? If the response to either of the questions is “Yes” then mark the document as belonging to this category. This category also includes short stories that may be published on an online platform.
\paragraph{Blogs}
A blog (short for “weblog”) is an online journal or informational website run by an individual, group, or corporation that offers regularly updated content (blog post) about a topic. It presents information in reverse chronological order and it's written in an informal or conversational style. You may have to look at the URL to check for this category. A blog typically has a title and addresses one topic throughout the text. Blogging has a highly personal form of writing and authors demonstrate a connection with their blog content.
\paragraph{Analytical Exposition}
The social function of Analytical Exposition text is: To persuade the reader that there is an important and correct matter that, certainly, needs to get attention. Analytical exposition typically uses emotive words and simple present tense. This type of text contains ads for products, properties, items, companies etc. It may even be in the form of a blog persuading the reader to either buy a certain product or avail certain services. In such situations the text should be first marked as a “Blog” and then as “Analytical Exposition”. This category includes persuasion, ads, and propaganda (text which is trying to sell the reader something or some idea).
\paragraph{Explanatory Article}
An explanatory article is a type of academic paper in which the author presents some point of view or opinion on a particular topic, subject, event or situation. Importantly, most of these articles provide references to the information presented in the text. This category includes Wikipedia articles, academic papers, abstracts of papers, Wiki How To articles or any piece of text plainly giving information for educational purposes. Note that any text that gives information is not an Explanatory Article. For example, in most cases ads also give information about a product but these should not be marked as Explanatory Articles. The purpose of Explanatory Articles is not to give information for selling something. These articles are also not written in conversation or informal format. They are written in a professional style and their sole purpose is to give information.
\paragraph{Reviews}
A review is a formal assessment or examination of something with the possibility or intention of instituting change if necessary. It is a critical article or report on a book, play, recital, movie, or an e-commerce product. A review typically provides a summary of the thing it is assessing, a reaction of the author and importantly a critical assessment of the thing.
\paragraph{Product/Company/Organization/Personal Websites}
Text that gives information about a product, company or organization falls into this category. The important thing is text in this category is authored and published by the same entity about which the information is given. For example a product website gives information about that product but a review website is written by someone else and will provide more than just the information about the product. Examples of this category are articles such as government websites giving information about their various programmes, organizations giving information about their services or products, schools giving information about courses, programmes, how to apply, jobs that are available etc.
\paragraph{Boilerplate Content}
Any written text (copy) that can be reused in new contexts or applications without significant changes to the original. Text and links in headers, footers, or sidebars are well-known examples. It could also be statements like “No search result” or email ids and addresses at the end of a website. Common examples of boilerplate are things like GDPR info about “cookies”, “Google analytics” for websites. Things like “about info” at the bottom of websites etc. If there are any HTML artifacts remaining in the article, this should be marked as boilerplate. Examples of HTML artifacts are things like tables <br>, <tr>, <html>. Oftentimes, javascript needed to render the web page can be embedded into the text, this should also be marked as boilerplate.
\paragraph{Miscellaneous}
Other categories not covered here so far. If the text contains pornographic content, or toxic / lewd / profane language then by default you should mark it as “MISC”.

We train a DeBERTaV3 model on this training set and find that on a held-out test set of 23k additionally labeled examples, it achieves an accuracy of 79.5\%. We plan to publicly release our type of speech classifier shortly.

% categories from doc
\section{Model Specifications}
\label{sec:appendix_model_specifications}
We detail the architecture and hyperparameters used for both the 2B and 8B models.

\paragraph{2B Model}

The architectural specifications include: 24 transformer layers, a hidden size of 2048, 16 attention heads, Rotary Position Embeddings (RoPE) \citep{su2023roformer}, SwiGLU \citep{shazeer2020glu} activations in the MLP layers, a SentencePiece \citep{kudo2018sentencepiece} tokenizer with a vocabulary size of 256k, a context length of 4096, no bias terms, and untied input-output embeddings.

We train with a batch size of 256 and use a cosine learning rate schedule, with warmup over the first one percent of training tokens, to decay from a maximum learning rate of $2.0e\text{-}4$ to  $2.0e\text{-}5$. We used the AdamW \citep{loshchilov2019decoupled} optimizer with $\beta_1 = 0.9$, $\beta_2 = 0.95$, and a weight decay of 0.1.

\paragraph{8B Model}

The architectural specifications include: 32 transformer layers, a hidden size of 4096, 32 attention heads, Rotary Position Embeddings (RoPE) \citep{su2023roformer}, SwiGLU \citep{shazeer2020glu} activations in the MLP layers, a SentencePiece \citep{kudo2018sentencepiece} tokenizer with a vocabulary size of 256k, a context length of 4096, no bias terms, and untied input-output embeddings.

We train with a batch size of 1024 and use a cosine learning rate schedule, with warmup over the first one percent of training tokens, to decay from a maximum learning rate of $3.0e\text{-}4$ to  $3.0e\text{-}5$. We used the AdamW \citep{loshchilov2019decoupled} optimizer with $\beta_1 = 0.9$, $\beta_2 = 0.95$, and a weight decay of 0.1.

\section{Data Curation Ablations}
\label{sec:appendix_data_curation_ablations}

Table \ref{tab:code_curation_results} illustrates that our specified steps of data curation for source code significantly improves evaluation performance, highlighting that data curation is a key component for all types of data.

\begin{table}[!h]
%\centering
  \begin{tabular}{lcc}
   \toprule
    \textbf{Experiment} & \textbf{HumanEval}  & \textbf{MultiPL-E}      \\
    \midrule
    Raw source code &  16.5 & 15.9 \\
    Post quality filtering & 20.7 & 19.2  \\
    \bottomrule
  \end{tabular}
  \caption{Evaluation accuracies before and after data curation for our source code dataset. We train an 8B model for 150B tokens.}
  \label{tab:code_curation_results}
\end{table}

\begin{table*}[!h]
\centering
\adjustbox{max width=\textwidth}{%
  \begin{tabular}{lcccccc}
   \toprule
    \textbf{Experiment} & \textbf{LAMBADA}  & \textbf{ARC-easy}  & \textbf{Race-H}  & \textbf{PIQA}  & \textbf{Winogrande}  & \textbf{Hellaswag}      \\
    \midrule
    Raw text &  55.6 & 57.2 & 39.9 & 73.9 & 57.6 & 58.9 \\
    Post deduplication &  57.8 & 59.1 & 39.9 & 76.6 & 56.9 & 63.3  \\
    Post quality filtering & 58.3 & 60.2 & 41.0 & 75.4 & 58.7 & 63.5  \\
    \bottomrule
  \end{tabular}}
  \caption{Per-task evaluation accuracies of the experiments detailed in Table \ref{tab:curation_results}.}
  \label{tab:full_data_curation_results}
\end{table*}

\begin{table*}[!h]
\centering
  \begin{tabular}{lcccccc}
   \toprule
    \textbf{Experiment} & \textbf{LAMBADA}  & \textbf{ARC-easy}  & \textbf{Race-H}  & \textbf{PIQA}  & \textbf{Winogrande}  & \textbf{Hellaswag}      \\
    \midrule
    Random &  59.8 & 59.4 & 41.9 & 75.6 & 59.9 & 63.1 \\
    Recent-to-Old   & 57.8 & 59.1 & 39.9 & 76.6 & 56.9 & 63.3 \\
    Old-to-Recent  & 59.4 & 60.8 & 41.3 & 76.0 & 61.7 & 63.5 \\
    \bottomrule
  \end{tabular}
  \caption{Per-task evaluation accuracies of the experiments detailed in Table \ref{tab:data_dedup_results}.}
  \label{tab:full_data_dedup_results}
\end{table*}

\begin{table*}[!h]
\centering
\adjustbox{max width=\textwidth}{%
  \begin{tabular}{llcccccc}
   \toprule
    \textbf{Question} & \textbf{Experiment} & \textbf{LAMBADA}  & \textbf{ARC-easy}  & \textbf{Race-H}  & \textbf{PIQA}  & \textbf{Winogrande}  & \textbf{Hellaswag}      \\
    \midrule
    \multirow{2}{*}{Q1} & 
    Original CC & 51.3 & 53.6 & 37.1 & 73.6 & 54.3 & 55.9 \\
    &  DSIR CC   & 53.1 & 55.0 & 37.2 & 73.2 & 54.4 & 53.8 \\
    \midrule
    \multirow{2}{*}{Q2.1} \hspace*{-0.25cm}
    & Corpus DSIR   & 53.1 & 55.0 & 37.2 & 73.2 & 54.4 & 53.8 \\
    & Source DSIR & 51.5 & 54.0 & 37.5 & 73.5 & 56.7 & 55.9 \\
    \midrule
    \multirow{3}{*}{Q2.2} \hspace*{-0.25cm}
    & DSIR (80\%)   & 53.3 & 5.40 & 37.4 & 72.5 & 56.5 & 53.6 \\
    & DSIR (87.5\%)   & 53.5 & 53.1 & 37.9 & 72.0 & 55.0 & 54.0 \\
    & DSIR (95\%)  & 51.5 & 54.0 & 37.5 & 73.5 & 56.7 & 55.9 \\
    \bottomrule
  \end{tabular}}
  \caption{Per-task evaluation accuracies of the experiments detailed in Table \ref{tab:dsir_main}.}
  \label{tab:full_dsir_main_results}
\end{table*}

\begin{table*}[!h]
\centering
\adjustbox{max width=\textwidth}{%
  \begin{tabular}{lcccccc}
   \toprule
    \textbf{Target Set} & \textbf{LAMBADA}  & \textbf{ARC-easy}  & \textbf{Race-H}  & \textbf{PIQA}  & \textbf{Winogrande}  & \textbf{Hellaswag}    \\
    \midrule
    Wikipedia, Books & 51.5 & 54.0 & 37.5 & 73.5 & 56.7 & 55.9 \\
    Wikipedia, Books, arXiv, NIH   & 46.9 & 53.6 & 38.2 & 74.3 & 55.6 & 55.6 \\
    arXiv, NIH   & 47.2 & 54.2 & 36.3 & 73.9 & 56.5 & 55.3 \\
    \bottomrule
  \end{tabular}}
  \caption{Per-task evaluation accuracies of the experiments detailed in Table \ref{tab:dsir_target_set_ablation}.}
  \label{tab:full_dsir_target_set_ablation}
\end{table*}

\section{Data Sampling Ablations}
\label{sec:appendix_data_sampling_ablations}

\begin{figure*}[h]
    \centering
    \includegraphics[width=\linewidth]{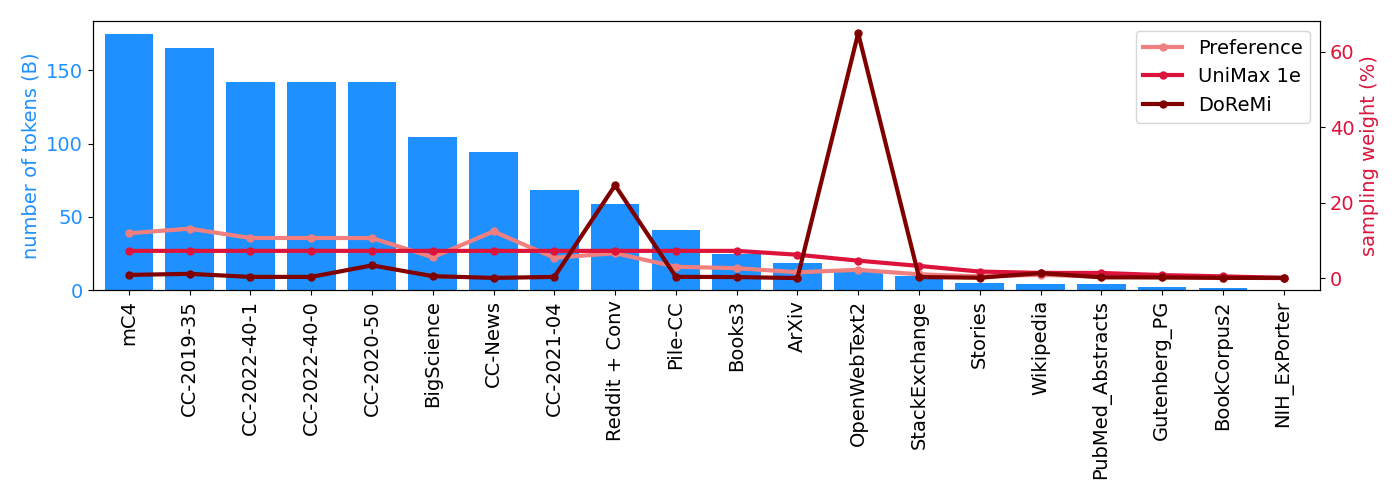}
    \caption{Returned samplings weights for the English dataset.}
    \label{fig:english_weights}
\end{figure*}

\begin{figure*}[h]
    \centering
    \includegraphics[width=\linewidth]{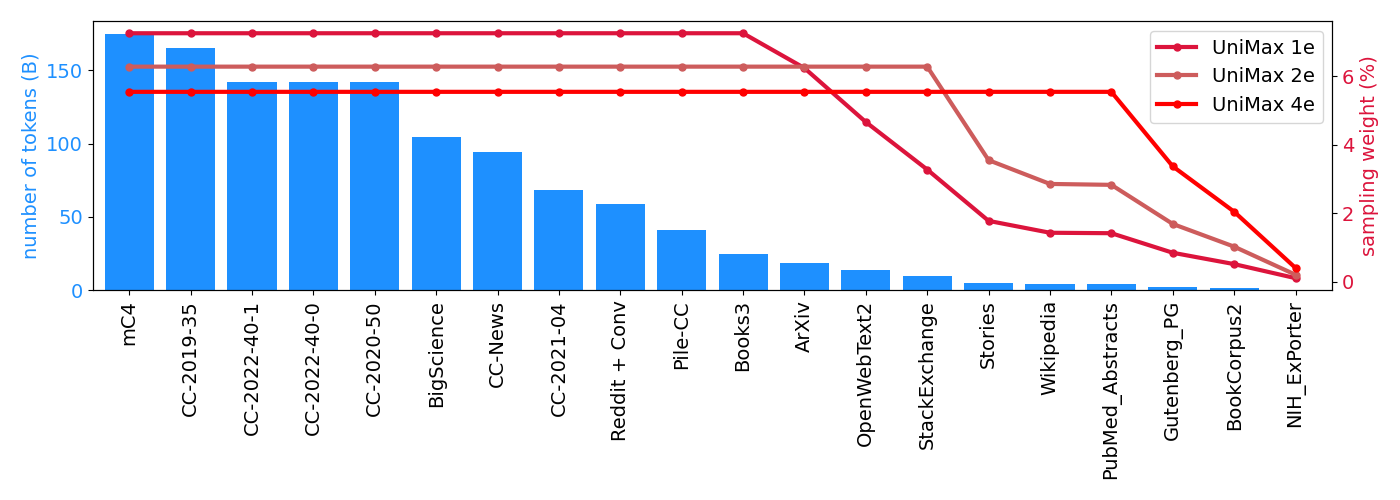}
    \caption{Effect of increasing the maximum epoch hyperparamter in UniMax on the returned sampling weights.}
    \label{fig:english_weights_uni}
\end{figure*}

\paragraph{English} We share the returned sampling weights for our English dataset across the three methods in Figure \ref{fig:english_weights} and across the varying values of the UniMax maximum epoch hyperparameter in Figure \ref{fig:english_weights_uni}. We clearly see that the returned weight distribution by DoReMi places too high of a weight on a single data source, which likely leads to its poor performance. Additionally, as the maximum epoch hyperparameter is increased in UniMax, the sampling distribution tends to a uniform one which likely begins to mitigate some of the utility gained from using the method.

\begin{table*}[!h]
\centering
\adjustbox{max width=\textwidth}{%
  \begin{tabular}{lccccccc}
   \toprule
    \textbf{Method} & \textbf{LAMBADA}  & \textbf{ARC-easy}  & \textbf{Race-H}  & \textbf{PIQA}  & \textbf{Winogrande}  & \textbf{Hellaswag} & \textbf{MMLU} \\
    \midrule
    Preference &  67.7 & 68.6 & 42.11 & 79.2 & 66.0 & 72.6 &  27.2 \\
    UniMax 1e  & 70.1 & 69.8 & 42.8 & 79.1 & 68.0 & 73.1 & 28.3  \\ 
    UniMax 2e   &70.7 & 67.6 & 42.9 & 78.9 & 66.3 & 72.6 & 28  \\ 
    UniMax 4e   & 70.5 & 67.7 & 43.0 & 78.9 & 67.3 & 72.4 &  26.6 \\ 
    DoReMi   & 68.3 & 68.6 & 41.2 & 78.9 & 65.0 & 72.0 & 26.9  \\ 
    \bottomrule
  \end{tabular}}
  \caption{Per-task evaluation accuracies of the experiments shared in \ref{tab:data_samp_eng}.}
  \label{tab:full_data_samp_eng}
\end{table*}

\begin{figure*}[h]
    \centering
    \includegraphics[width=\linewidth]{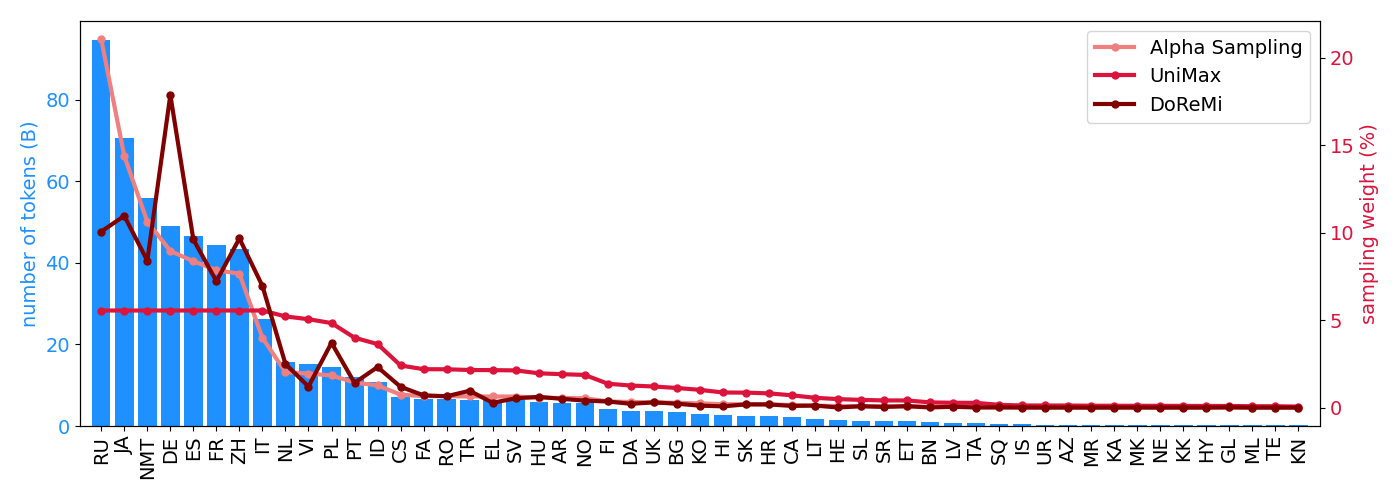}
    \caption{Returned samplings weights for the Multilingual dataset.}
    \label{fig:multilingual_weights}
\end{figure*}

\paragraph{Multilingual} In our multilingual ablations, we first ran a series of experiments to identify the optimal $\alpha$ value to use in alpha sampling. We found that $\alpha=1.3$ achieved the best downstream accuracies. We share the returned sampling distribution from each method in Figure \ref{fig:multilingual_weights}.

\begin{figure*}[h]
    \centering
    \includegraphics[width=\linewidth]{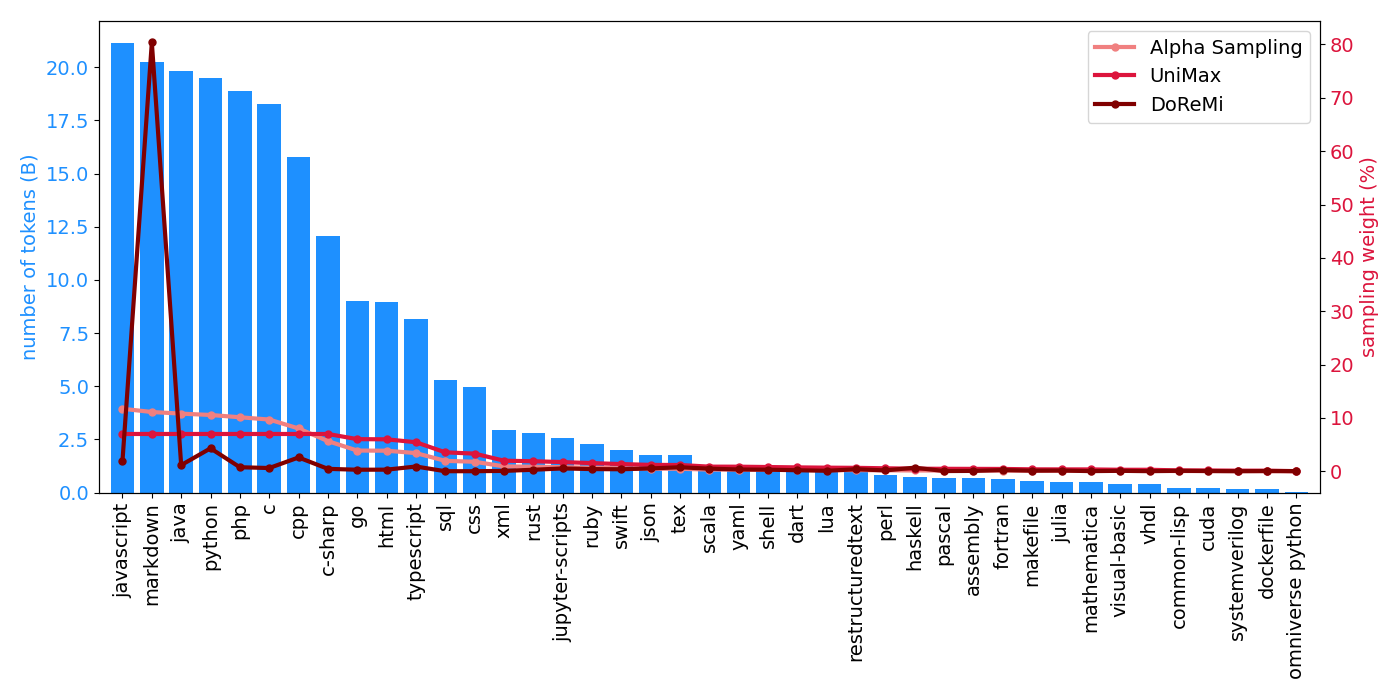}
    \caption{Returned samplings weights for the Code dataset.}
    \label{fig:code_weights}
\end{figure*}

\paragraph{Code} Like in our multilingual ablations, we found that  $\alpha=1.3$ achieved the best downstream accuracies for alpha sampling in the code domain. We share the returned sampling distribution from each method in Figure \ref{fig:code_weights}. The DoReMi identified sampling distribution is not useful as it places over 80\% of the weight on markdown.

\begin{table*}[!h]
\centering
  \begin{tabular}{lcccccc}
   \toprule
      \textbf{Method} & \textbf{HumanEval} & \textbf{MP-Python} & \textbf{MP-Java} & \textbf{MP-JS} & \textbf{MP-CPP} & \textbf{MP-Lua} \\
    \midrule
    Alpha   &  20.72 & 20.5 & 23.4 & 20.5 & 19.3 & 14.5 \\
     UniMax   & 20.12 & 19.3 & 20.9 & 19.9 & 19.3 & 17.4 \\
    \bottomrule
  \end{tabular}
  \caption{Per-task evaluation accuracies for the experiments detailed in Table \ref{tab:data_samp_code}. MP stands for MultiPL-E.}
  \label{tab:full_data_samp_code}
\end{table*}

\section{Data Attribute Analysis}
\label{sec:appendix_data_attribute_analysis}

\begin{figure}[h]
    \centering
    \includegraphics[width=\linewidth]{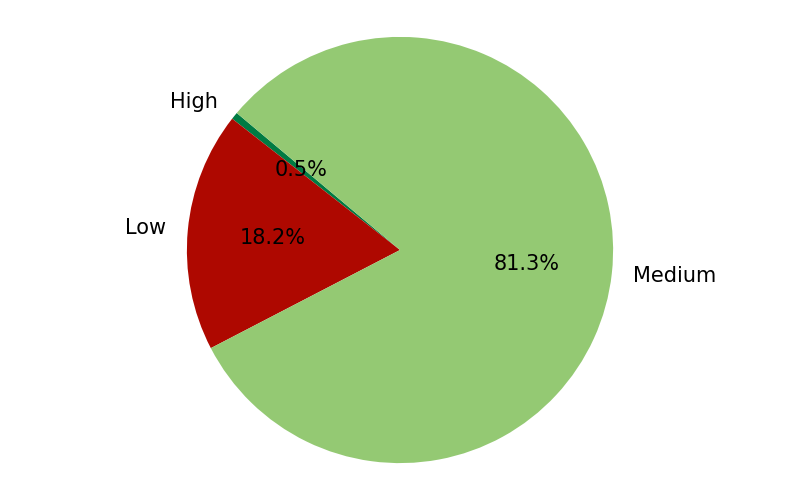}
    \caption{Breakdown of document quality across web crawl snapshots.}
    \label{fig:quality}
\end{figure}

\begin{figure}[h]
    \centering
    \includegraphics[width=\linewidth]{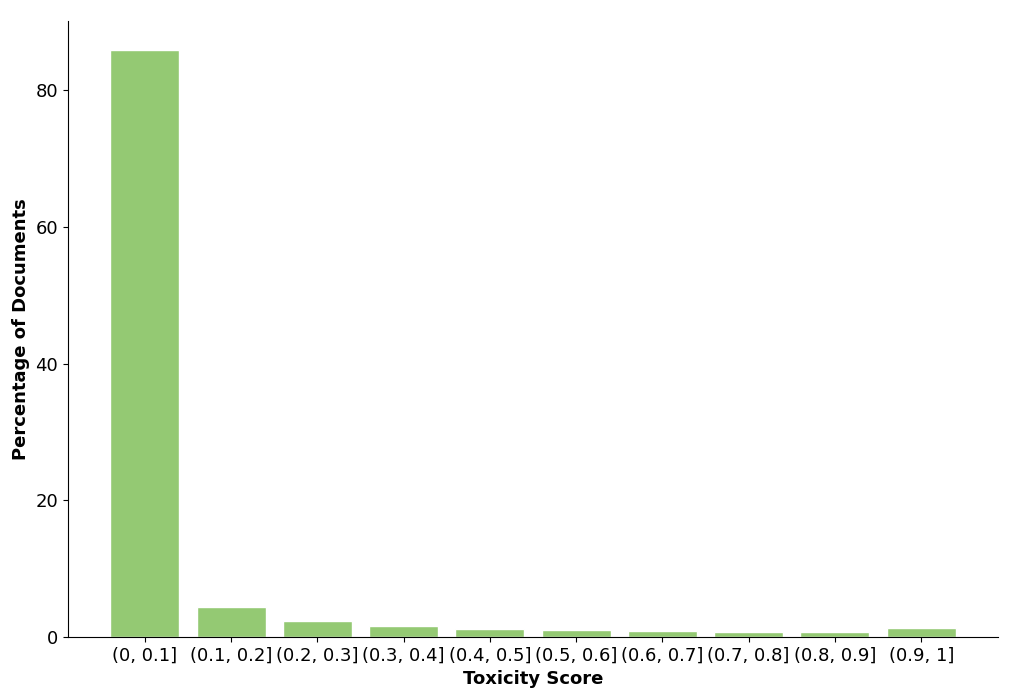}
    \caption{Breakdown of document toxicity across web crawl snapshots.}
    \label{fig:toxicity}
\end{figure}

 Figure \ref{fig:quality} illustrates that the vast majority of web crawl documents are of medium quality; however, there does exist a significant chunk of low quality documents which should be appropriately considered when creating pretraining sets. Additionally, Figure \ref{fig:toxicity} highlights that a large proportion of web crawl documents are unlikely to contain toxic content (defined as having a toxicity score lower than 0.3). These two factors combined assure us that web crawl snapshots provide positive utility during langauge model pretraining.

\begin{figure}[h]
    \centering
    \includegraphics[width=\linewidth]{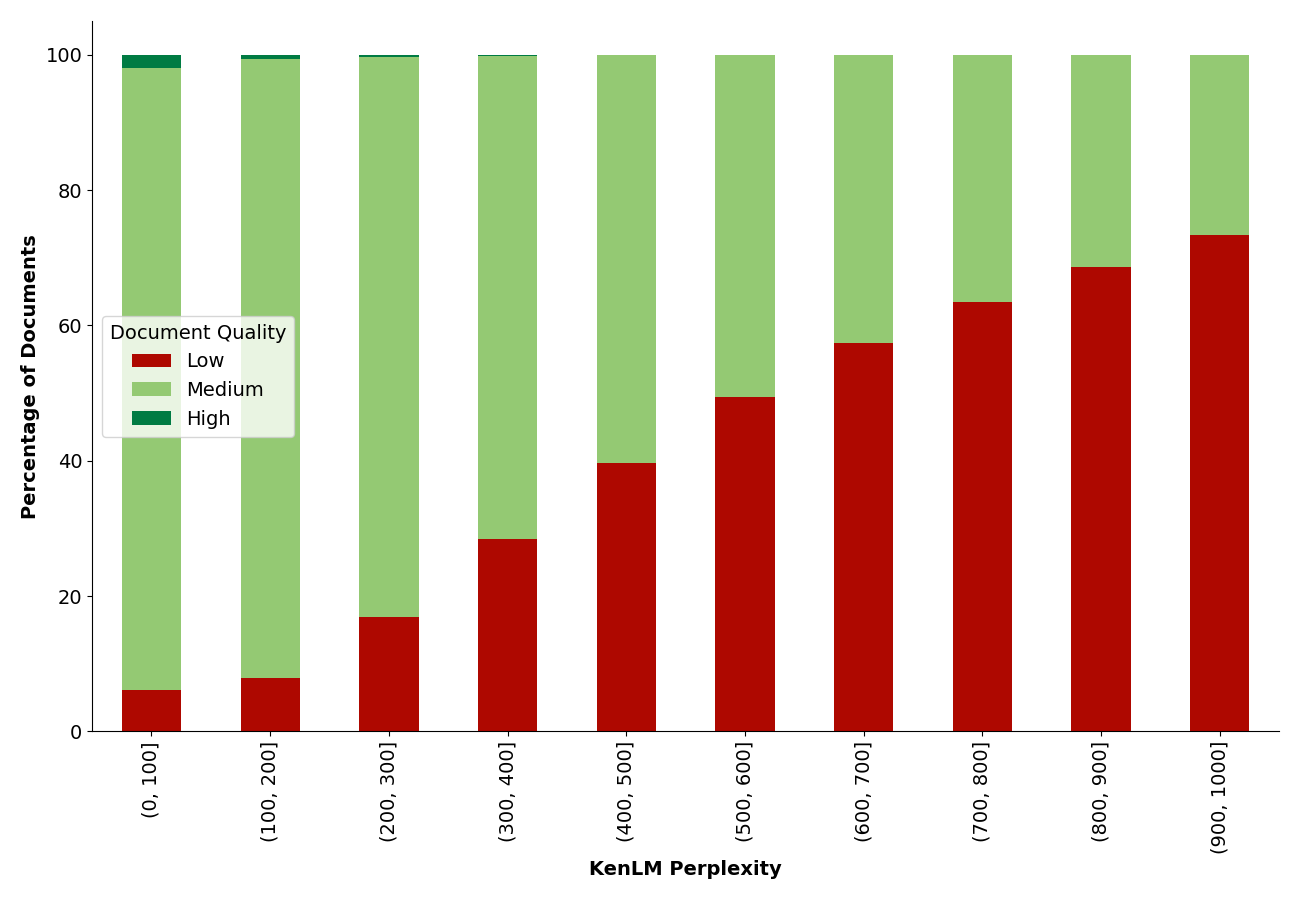}
    \caption{There is high correlation between the quality classifier and the perplexity of a KenLM model used for quality filtering during data curation.}
    \label{fig:quality_kenlm}
\end{figure}

Next, we examine the overlap between the output of the developed quality classifier and the perplexity scores of the KenLM model which we used to filter low quality documents during data curation. Figure \ref{fig:quality_kenlm} shows that the two models have high agreement on documents which they classify as high or low quality. This indicates that such model based filtering during data curation is able to reliably remove low quality texts.  

\begin{figure}[h]
    \centering
    \includegraphics[width=\linewidth]{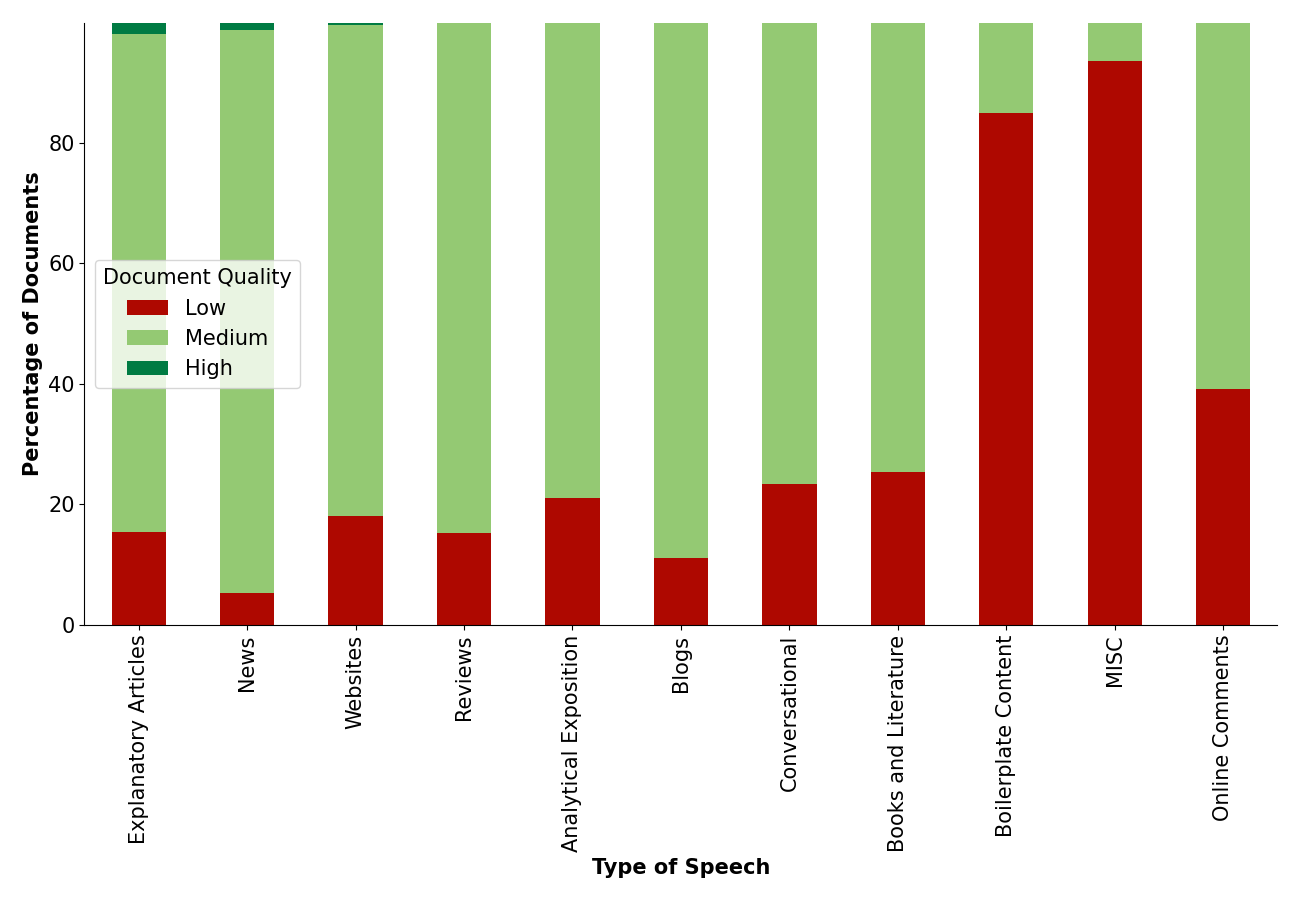}
    \caption{Types of speech sorted by descending order of percentage of high quality documents.}
    \label{fig:tos_qual}
\end{figure}

%% HERE
In examining the quality composition of various types of speech categories, as shown in Figure \ref{fig:tos_qual}, we find that explanatory and news articles are the document types which tend to contain the highest proportion of high quality texts. Additionally, we see that the boilerplate content and miscellaneous categories by far have the largest proportion of low quality documents, indicating that it likely would be best to completely filter out web domains which contain high proportions of documents of these types. This analysis allows for the appropriate prioritization of document types within web crawl snapshots as we now understand which sorts of texts are likely to be of the highest quality.

\begin{figure}[h]
    \centering
    \includegraphics[width=\linewidth]{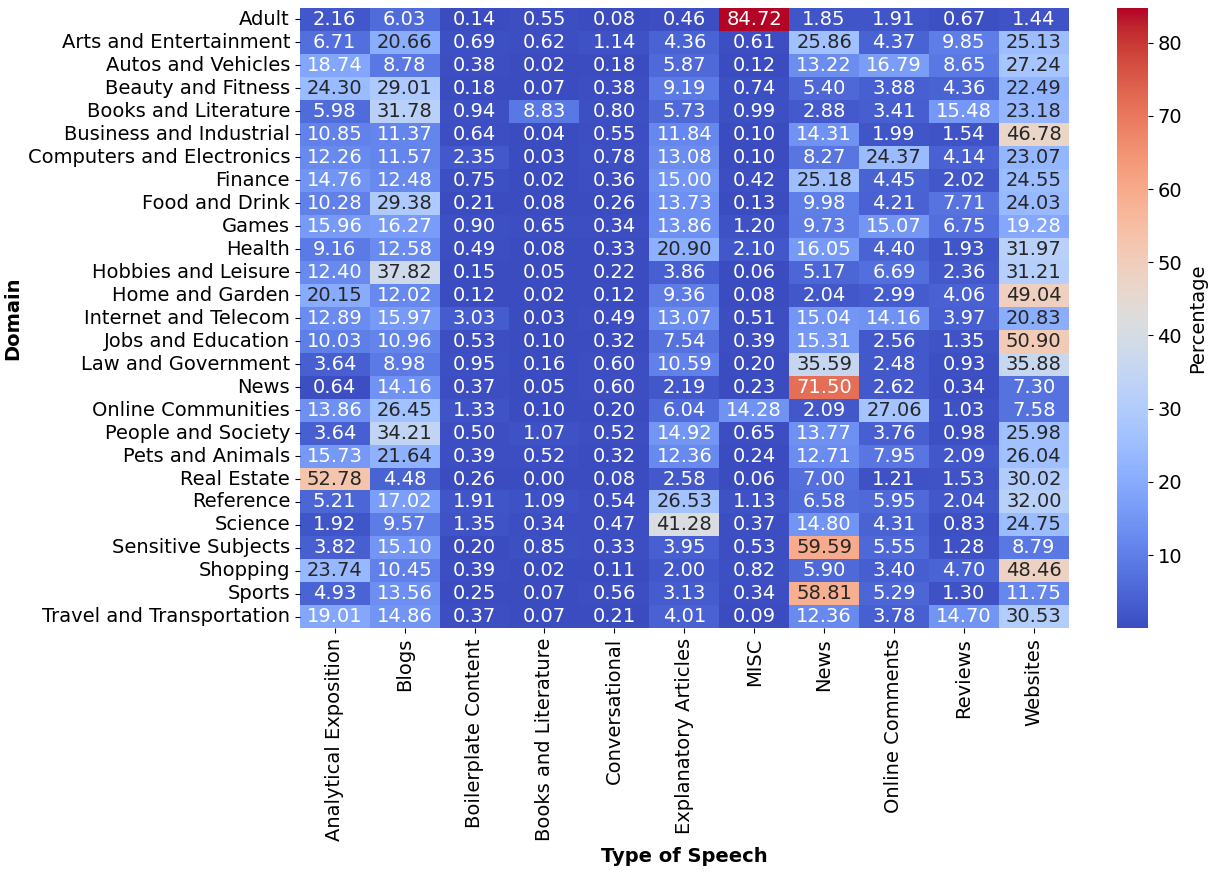}
    \caption{Heatmap of domains by types of speech.}
    \label{fig:domains_tos}
\end{figure}

Lastly, Figure \ref{fig:domains_tos} highlights the distribution of domain by type of speech. We find that a lot of the technical domains, such as science, law, and health, are primarily composed of high quality types of speech, such as news and explanatory articles. This highlights that when prioritizing certain websites in future web crawls, it likely would be most fruitful to focus on those surrounding such domains. Additionally, the domain of sensitive subjects, which we identified as being primarily composed of high quality documents, is in fact made up mostly by news articles. This would indicate that this domain likely covers investigative reports on subjects such as war and protests. We also note that the categories which we expect to have high overlap, like the domain and type of speech of news or the adult domain and the miscellaneous type of speech category, do in fact have a high degree of overlap. This confirms the efficacy of both our classifiers in providing accurate analysis.

\section{Data Attributes in Sampling and Selection}
\label{sec:appendix_data_attribute_experiments}

In this set of experiments, our baseline data sampling method is to proportionally weight each of the 5 CC snapshots by their token counts. We found that this sampling method performed better than UniMax. As the CC snapshots are all of relatively large token counts compared to our training token budget, 165B, UniMax ends up assigning a uniform distribution across each of the snapshots. As different CC snapshots have different utility, as indicated by \citep{fine-web}, a uniform distribution is suboptimal to one which weights snapshots differently. 

In defining the sampling weights over both the \texttt{Fine-Grained} and \texttt{Grouped} settings of the attribute based buckets, we use UniMax with the maximum epoch hyperparameter set to 2.

\begin{table*}[!h]
\adjustbox{max width=\textwidth}{%
\centering
  \begin{tabular}{lccccccc}
   \toprule
      \textbf{Experiment} & \textbf{LAMBADA}  & \textbf{ARC-easy}  & \textbf{Race-H}  & \textbf{PIQA}  & \textbf{Winogrande}  & \textbf{Hellaswag}   \\
     \midrule
     Baseline & 54.1 & 56.5 & 38.9 & 75.1 & 57.8 & 58.9 \\
     \midrule
     Quality \texttt{Fine-Grained}  & 57.3 & 57.7 & 39.7 & 75.0 & 57.6 & 60.0 \\
     Quality \texttt{Grouped}  & 56.2 & 56.6 & 38.7 & 74.2 & 56.8 & 58.3 \\
     \midrule
     Toxicity \texttt{Fine-Grained}  & 46.1 & 57.6 & 36.9 & 71.3 & 55.5 & 46.2 \\
     Toxicity \texttt{Grouped}  & 55.0 & 56.1 & 37.3 & 72.7 & 54.5 & 54.2 \\
     \midrule
     Domain \texttt{Fine-Grained}  & 57.0 & 60.7 & 39.5 & 73.3 & 56.5 & 57.0 \\
     Domain \texttt{Grouped}  & 54.6 & 59.7 & 40.2 & 73.9 & 59.2 & 57.1 \\
     \midrule
     Type of Speech \texttt{Fine-Grained}  & 53.4 & 59.2 & 37.5 & 74.3 & 56.2 & 59.5 \\
     Type of Speech \texttt{Grouped}  & 53.9 & 59.8 & 37.5 & 74.3 & 58.7 & 59.6 \\
    \bottomrule
  \end{tabular}}
  \caption{Per-task evaluation accuracies of the experiments detailed in \ref{tab:data_attr_samp}.}
  \label{tab:full_data_attr_samp}
\end{table*}

%\section{Case Study}
%\label{sec:appendix_case_study}

\end{document}